\documentclass[conference]{IEEEtran}
\usepackage{cite}
\usepackage{amsmath,amssymb,amsfonts}
\usepackage{enumerate}
\usepackage{graphicx}
\usepackage{subfigure}
\usepackage{textcomp}
\usepackage{xcolor}
\usepackage{makecell}
\usepackage{stfloats}
\usepackage{gensymb}
\usepackage{multirow}
\usepackage{soul, color}
\usepackage{hyperref}
\usepackage{cleveref}
\makeatletter
\renewcommand{\@thesubfigure}{\hskip\subfiglabelskip}
\makeatother

\def\hrulefill{\leavevmode\leaders\hrule height 0.8pt\hfill\kern0pt}
\makeatletter
\def\rulefill{\leavevmode\leaders\hrule depth -3pt height 4pt\hfill\kern0pt}
\makeatother
\def\BibTeX{{\rm B\kern-.05em{\sc i\kern-.025em b}\kern-.08em
    T\kern-.1667em\lower.7ex\hbox{E}\kern-.125emX}}
\begin{document}



\title{Automatic Localization and Detection Applicable to Robust Image Watermarking Resisting against Camera Shooting}

\author
{\IEEEauthorblockN{Ming Liu$^1$, Si Li$^{*2}$,Wu Wang$^1$}
	\IEEEauthorblockA{
		\text{liuming@gznu.edu.cn}
		}}
\maketitle

\begin{abstract}
Robust image watermarking that can resist camera shooting has become an active research topic in recent years due to the increasing demand for preventing sensitive information displayed on computer screens from being captured. However, many mainstream schemes require human assistance during the watermark detection process and cannot adapt to scenarios that require processing a large number of images. Although deep learning-based schemes enable end-to-end watermark embedding and detection, their limited generalization ability makes them vulnerable to failure in complex scenarios. In this paper, we propose a carefully crafted watermarking system that can resist camera shooting. The proposed scheme deals with two important problems: automatic watermark localization (AWL) and automatic watermark detection (AWD). AWL automatically identifies the region of interest (RoI), which contains watermark information, in the camera-shooting image by analyzing the local statistical characteristics. Meanwhile, AWD extracts the hidden watermark from the identified RoI after applying perspective correction. Compared with previous works, the proposed scheme is fully automatic, making it ideal for application scenarios. Furthermore, the proposed scheme is not limited to any specific watermark embedding strategy, allowing for improvements in the watermark embedding and extraction procedure. Extensive experimental results and analysis show that the embedded watermark can be automatically and reliably extracted from the camera-shooting image in different scenarios, demonstrating the superiority and applicability of the proposed approach.
\end{abstract}


\section{Introduction}
\begin{figure}[!htbp]
	\centering
	\subfigure[(a) Screen-shooting scenarios]{\includegraphics[height=1.5 in]{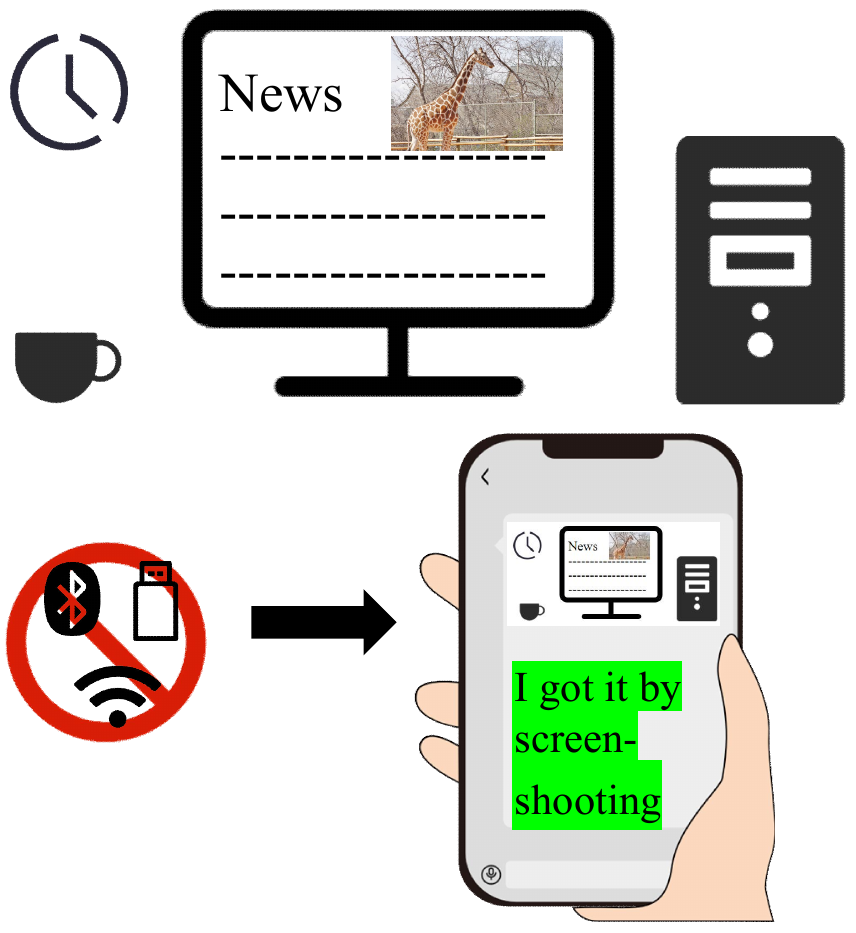}}
	\hspace{5 pt}
	\subfigure[(b) Private image leaking]{\includegraphics[height=1.5 in]{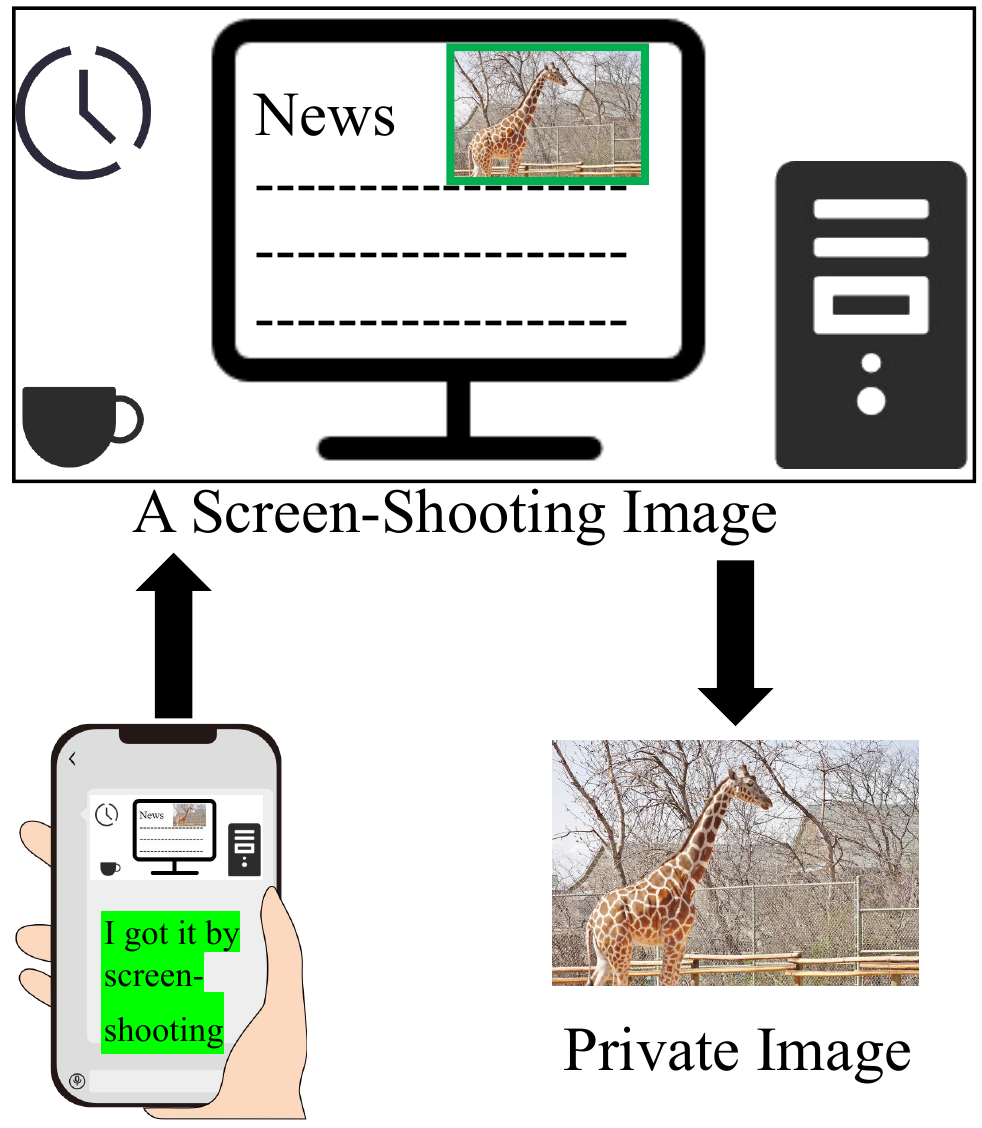}}
	\caption{Screen-shooting scenarios and private image leaking problem.}
	\label{fig:FF}
\end{figure}

\begin{figure*}[!htbp]
	\centering
	\includegraphics[width=5 in]{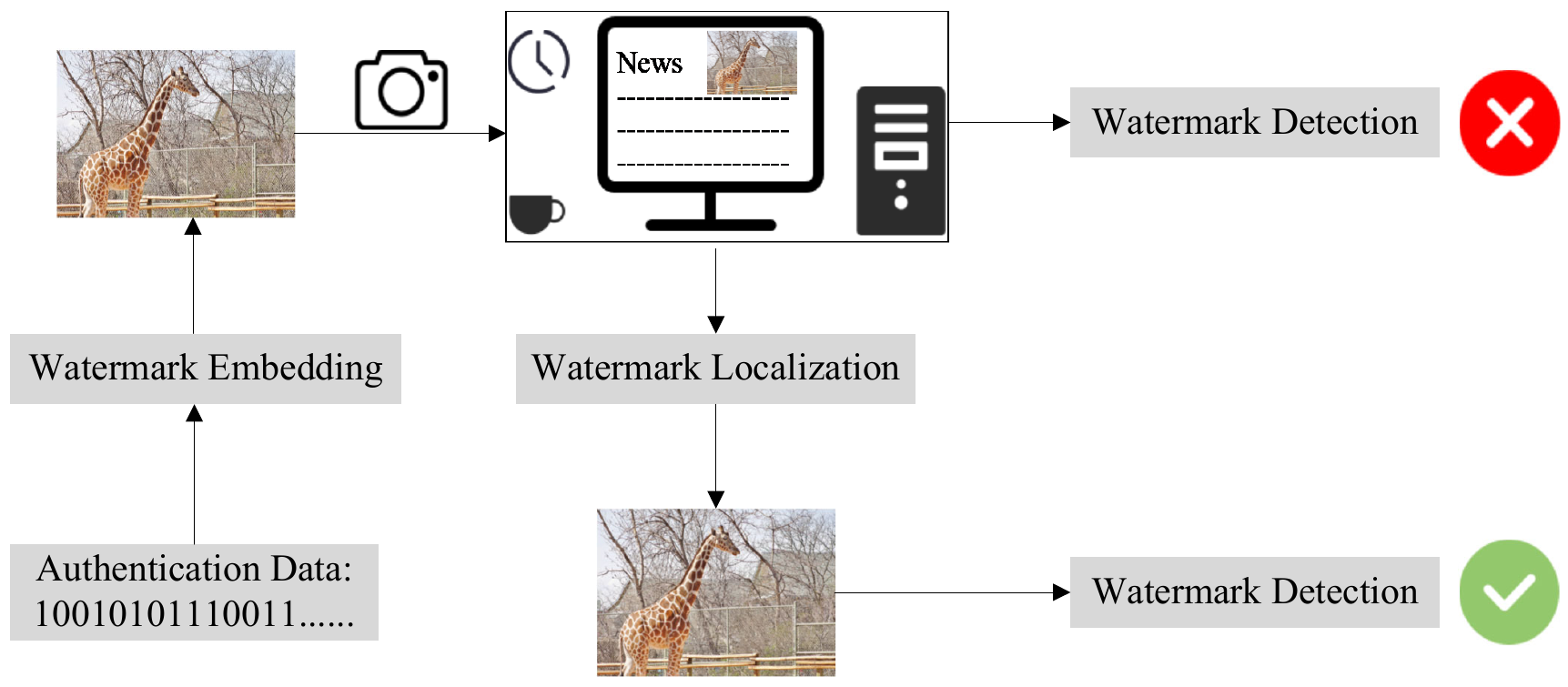}
	\caption{A framework for watermark embedding, localization and detection in screen-shootings scenarios.}
	\label{fig:intro}
\end{figure*}

In recent years, the pervasive preference for communicating information through photographs has emerged due to the widespread use of mobile phones. However, this convenience has also led to the perilous outcome of confidential content being disclosed, as illustrated in Fig.~\ref{fig:FF}.
For example, an image for private purposes, such as the image of giraffe in Fig.1, can be recklessly exposed to the public sphere. To prevent unauthorized dissemination of private images as mentioned above, watermarking has proven to be an effective solution.
It is worth noting that, in the instant case, the image of a giraffe, which is inherently intended for private purposes, has been recklessly exposed to the public sphere.
In order to mitigate the aforementioned issue of unauthorized dissemination of private images, watermarking has proven to be a remarkably efficacious solution. To expound upon this point, it should be mentioned that the majority of extant methodologies for addressing this issue involve the embedding of authentication data into the private image, which can subsequently be traced via the detection of watermarks. However, some watermark localization methodologies may pose challenges in practical implementation. For instance, the proposed schemes in references \cite{fang2018screen,LiDong2022WATERMARKPRESERVINGKE,chen2020screen,deng2022svd}  require human assistance, while in \cite{gourrame2022fourier} a private image can only be located under ideal shooting conditions by detecting corners;in \cite{wang2022print}, the original image is required to register the screen shooting image (SSI) based on paired feature point.
In screen-shooting scenarios, accurate watermark localization is of paramount importance as it facilitates subsequent detection of the watermark. As depicted in Fig.~\ref{fig:intro} aptly illustrates, failure to precisely pinpoint the location of the private image can result in the regrettable outcome of the embedded authentication data becoming irretrievably lost. Moreover, existing localization methodologies that need human intervention are impractical for applications involving a large number of images. The credibility of forensics involving manual operations is also invariably subject to reasonable doubt.


In the past decades, digital watermarking \cite{van1994digital,hsu1999hidden,boney1996digital} was referred to as the art of embedding secret information within a cover such as image, audio and video to achieve copyright protection\cite{abdelhakim2017quality,PeymanAyubi2021ANC} or active forensics\cite{chen2005fragile,PascalLefvre2022EfficientIT,raj2021survey}. In scenarios requiring copyright protection, the watermarking method needs to be robust enough to withstand various digital signal processing and geometric transformation attacks, ensuring that the hidden information can survive from those attacks\cite{ma2021local,huan2021exploring}. For active forensics scenarios, the digital watermarking must be fragile, making it easy to be destroyed by human malicious attacks such as copy-move, removal and addition. In these cases the accuracy was vital for fragile digital watermarking\cite{d2018patchmatch,AdityaKumarSahu2021ALM}.

Recently, the issue for copyright infringement due to screen-shooting has attracted widespread attentions \cite{fang2018screen,chen2020screen,LiDong2022WATERMARKPRESERVINGKE}. Most methods typically embed authentication data in the private image and extract authentication data from the SSI. However, screen-shooting can cause serious de-synchronization attacks, e.g. location shifting, shape transforming, and scaling, which lead to error extraction. To wrestle with these problems, most existing methods consider the de-synchronization attacks as regular tampering and locate the screen-shooting private images by manual operations. Furthermore, end-to-end convolutional neural network (CNN) based methods have been used to solve this problem with labeled dataset \cite{jia2020rihoop,wengrowski2019light,fang2022pimog}, but its accuracy is limited by the size of dataset.

In recent years, robust watermarking methods for screen-shooting scenarios have been studied and can be classified into three categories: feature points based methods\cite{alghoniemy2004geometric}, template based methods\cite{pereira2000robust,kang2003dwt} and deep learning based methods\cite{tancik2020stegastamp,wengrowski2019light,jia2020rihoop}.

For feature points based methods, Fang et al. \cite{fang2018screen} proposed an intensity-based scale-invariant feature transform (I-SIFT) algorithm\cite{lowe2004distinctive} which can accurately locate the embedding regions. They put forward a small-size template algorithm to repeatedly embed watermarks into different regions, ensuring at least one complete information region can survive. Chen et al. \cite{chen2020screen} proposed a feature-based synchronization algorithm based on speeded-up robust feature (SURF) orientation descriptor\cite{bay2006surf}. The message is repeatedly embedded in each selected local square feature region (LSFR). They employed a non-rotating embedding method and a preprocessing method to modulate the discrete Fourier transform (DFT) coefficients. Dong et al. \cite{LiDong2022WATERMARKPRESERVINGKE} proposed a method that jointly optimized the watermarking distortion and enhanced keypoints. Their proposed method achieved superior watermark extraction accuracy while retaining better watermarked image quality compared than Fang's work \cite{fang2018screen}. Chang et al. \cite{chang2022blind} proposed a novel blind video watermarking scheme to improve the robustness by introducing oriented fast and rotated brief (ORB) \cite{rublee2011orb} key points to select regions against geometric attacks and camera recording attacks.

For template based methods, NaKamura et al. utilized a set of orthogonal templates to represent the 0/1 bit and add the corresponding templates to the cover image to embed the watermarks. The extraction process involved matching templates \cite{nakamura2004fast}. Kim et al. proposed a method to
embed messages in the form of pseudo-random vectors. To extract the message, the autocorrelation function detects the grid formed by the message, followed by applying the cross-correlation operation to extract the message\cite{kim2006image}. Pramila et al. proposed a multi-scale template based method, in which they generated a periodic template and encoded the information by modulating the directions of templates. Then, the embedding capacity for a unit block is effectively expanded and the message is extracted by detecting the angle of templates using Hough transforming. \cite{pramila2009reading,pramila2012toward}. In the last few years, Pramila et al. also optimized the image processing procedure for this method, which effectively enhanced both the
visual quality of the image and the robustness of the watermark \cite{pramila2017extracting,pramila2018increasing}.

For many deep learning-based methods, an attack simulator\cite{zhu2018hidden} is often built to acquire feedback. A scheme of simulating the whole process of screen photography was proposed by Jia et al.\cite{jia2020rihoop}. They built a distortion network between encoder and decoder to augment the encoded images. The distortion network used differentiable 3-D rendering operations to simulate the distortion introduced by camera imaging in both printing and display scenarios. Another typical scheme was to model the camera-display pipeline \cite{wengrowski2019light}. In this method, Wengrowski et al. proposed a key component that was the camera display transfer function (CDTF) to model the camera-display pipeline. To learn this CDTF, they introduced a dataset called Camera-Display-1M, which contains 1,000,000 camera-captured images collected from 25 camera-display pairs. Recently, Fang et al.\cite{fang2022pimog} proposed an effective screen-shooting noise-layer simulation for deep-learning-based watermarking network. They summarized most influenced distortions of screen-shooting process into three parts (perspective distortion, illumination distortion and moiré distortion) and further simulated them in a differentiable way. For the other types of distortion, they utilized the Gaussian noise to approximate them.

Automatic watermark localization (AWL) in SSI is often overlooked. For the essential feature of screen-shooting, several challenges for  AWL are as follows:
\begin{itemize}
	\item \textbf{The presence of superfluous entities within a screen-shooting image can prove to be a major impediment to the detection technique.} The act of screen-shooting inevitable includes capturing extraneous objects in the image, such as clocks and cups as illustrated in Fig.~\ref{fig:FF}, which can erroneously construed as the intended targets and thus precipitate an erroneous detection. Additionally, identifying a designated region of interest (RoI) within an SSI using machine learning is particularly challenging when there is no prior knowledge or context available.

	\item \textbf{In screen-shooting scenarios, various attacks may occur, making it imperative to fortify watermarking against these onslaughts.} Screen-shooting process requires robust watermarking as it can resist adversarial blend attacks such as re-sampling, re-compression, and re-coloring. Furthermore, due to the unpredictable nature of perspective transformation applied to target images, rectangular bounding boxes may not be relied upon to facilitate RoI detection using existing methodologies.

	\item \textbf{Flexible scales are introduced by personalized record devices to make the size of embedding blocks changed and cause the incorrect detecting.} Personalized recording devices have engendered a novel mechanism that enables flexible scaling, allowing for the alteration of the size of embedding blocks, which may lead to a spate of spurious detections. This issue is further compounded by the fact that screen-shooting images are typically captured in a larger size than their digital counterparts, with the precise scale of the former contingent upon the particulars of the recording device.
\end{itemize} 

In light of the aforementioned hurdles, an innovative solution has been conceived: a cutting-edge automatic localization and detection scheme that is tailored to the needs of robust image watermarking in the context of camera shooting.
The contributions of this paper can be summarized as follows:
\begin{enumerate}
\item a universal solution has been proposed that is to detect regions of interest in screen-shooting images with unparalleled accuracy, thereby enabling the attainment of a precise target image location within screen-shootings images. Remarkably, the proposed scheme is not tethered to any particular watermark embedding strategy, and enabling us to enhance the watermark embedding and extraction methodologies without any unnecessary constraints.

\item An avant-garde and highly sophisticated two-stage non-maximum suppression scheme has been conceptualized with a view to pinpointing the target image with unparalleled precision. In the first stage, an intensity set is deployed to establish the optimal scale that should be retained for a specific SSI. The second stage is characterized by the application of a novel local cost method that enables the identification of the optimal intersect line.



\item The proposed scheme adapts well to various complex scenarios, including curved screens, re-shooting, Print-camera, print-crop-camera, and gray-scale printing, as demonstrated by experimental results.
\end{enumerate}
 

\section{Proposed method}
\begin{figure}[!htbp]
	\centering
	\includegraphics[width=3.5 in]{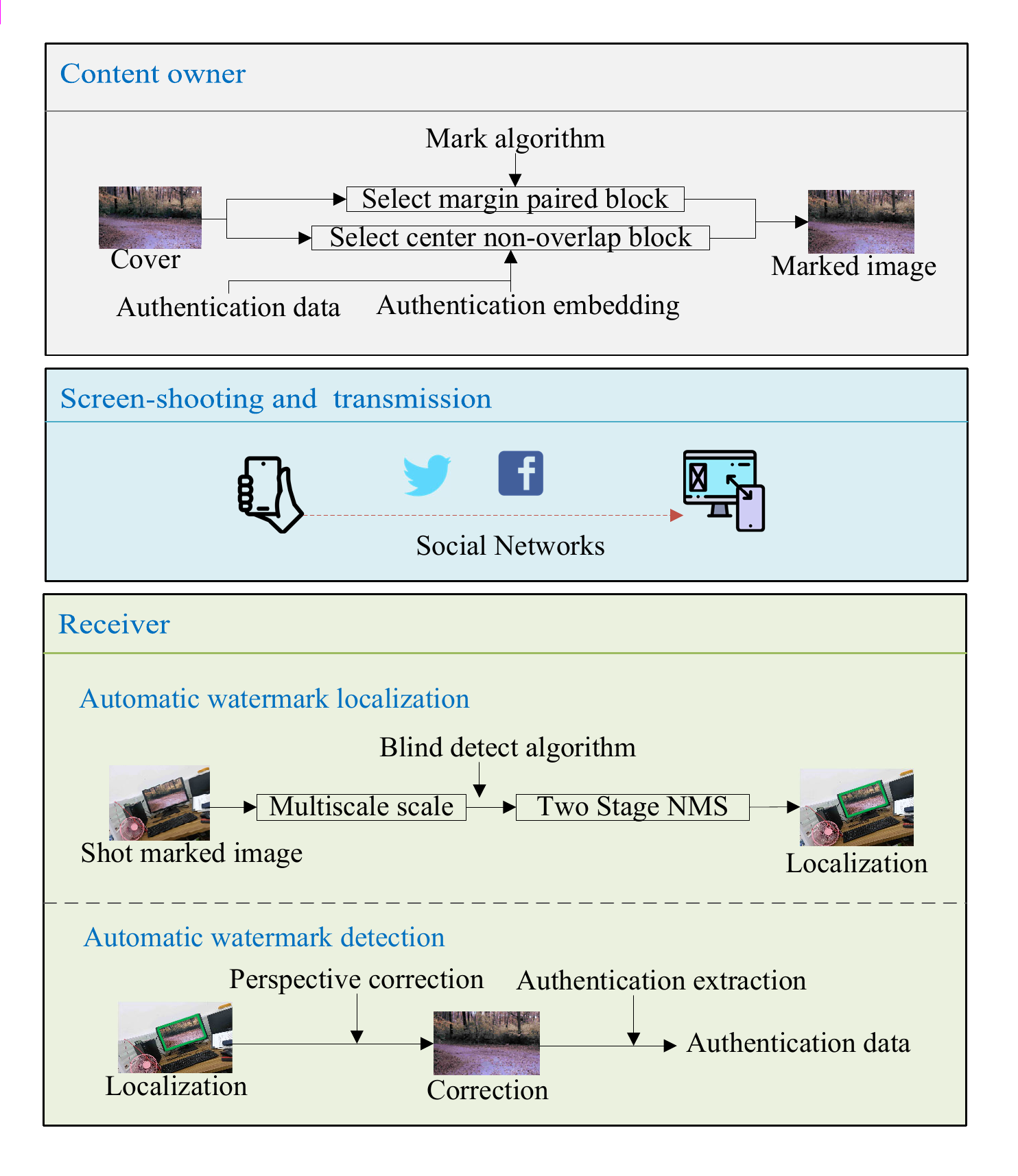}
	\caption{General framework for the proposed method.}
	\label{fig:overview}
\end{figure}

In this section, we shall present the underlying principles of our proposed automatic localization and detection scheme that can effectively counter the adverse effects of camera shooting on robust image watermarking. As depicted in Fig.~\ref{fig:overview}, our method consists of several key steps. Initially, we employ a marking algorithm to embed watermarks in the private image, ensuring that it can be detected in the corresponding SSI. In the subsequent localization and detection section, we employ multi-scale scale and a two-stage non-maximum suppression mechanism to identify the region of interest (RoI) in the SSI without any prior knowledge. Finally, we introduce a novel scheme of automatic watermarking detection that combines several typical watermarking algorithms to evaluate its efficacy in detecting target images based on watermarking.

\subsection{Marking algorithm}
Screen-shooting is a challenging task as it introduces various complexities, which make it difficult to detect the Region of Interest (RoI) in screen-shooting image (SSI). In the absence of any fixed or invariant features, detecting the RoI becomes even more complicated. Therefore, to address this issue, the most straightforward approach is to introduce handcrafted features. In this study, we propose to use watermarking to generate such features.

\begin{figure*}[!htbp]
	\centering
	\includegraphics[width=7 in]{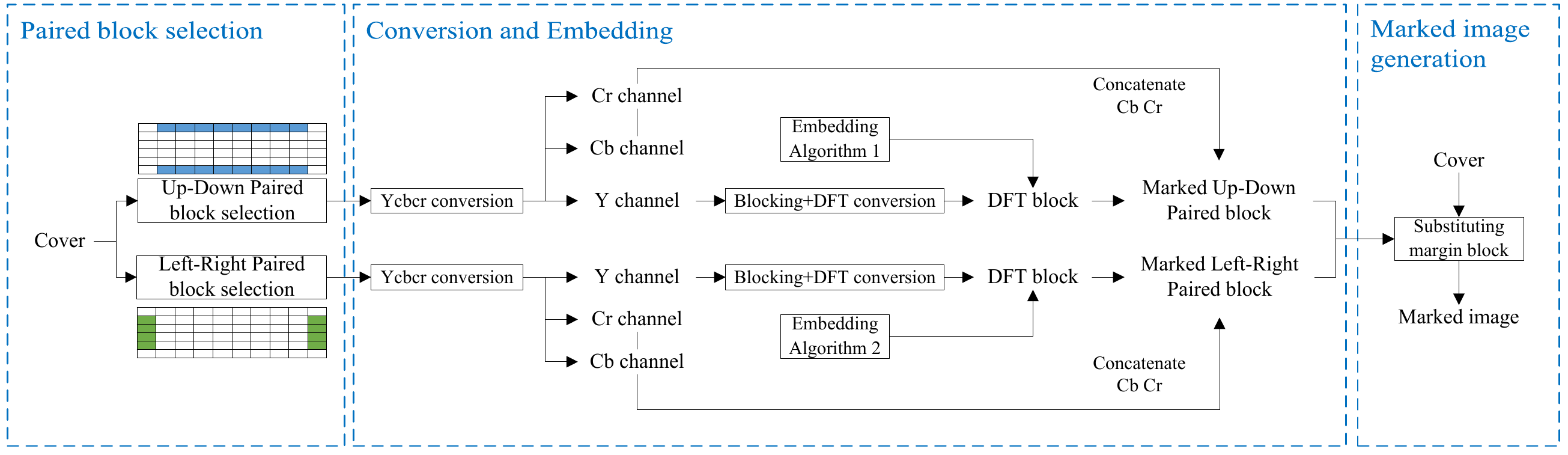}
	\caption{The framework of proposed marked algorithm}
	\label{fig:marked_framework}
\end{figure*}

To nail down the accurate location of RoI in SSI, we conduct paired blocks selection and orthogonal transformation before diving into the marking operation. And you can see the nitty-gritty details of this framework in Fig.~\ref{fig:marked_framework}.

\subsubsection{Paired blocks selection}
we select the up and down margin blocks as $GroupA=\{a_1,a_2,...,a_x\}$, then the left and right margin blocks as another $GroupB=\{b_1,b_2,...,b_y\}$, where $a_i$ and $b_j$ are non-overlap sub-blocks in each group respectively, and we assume the sub-blocks sizes as $l \times l$. $x>0,y>0$ are numbers of sub-blocks in $GroupA$ and $GroupB$ respectively. An example is given in Fig.~\ref{fig:marked_framework}.

\subsubsection{Conversion and Embedding}
\subsubsection*{Conversion}
we convert the $a_i$ or $b_j$ to YCbCr colorspace and then choose the Y-channel as the host block, then apply DFT conversion on them.
\subsubsection*{Embedding}
\begin{figure*}[!htbp]
	\centering
	\includegraphics[width=7 in]{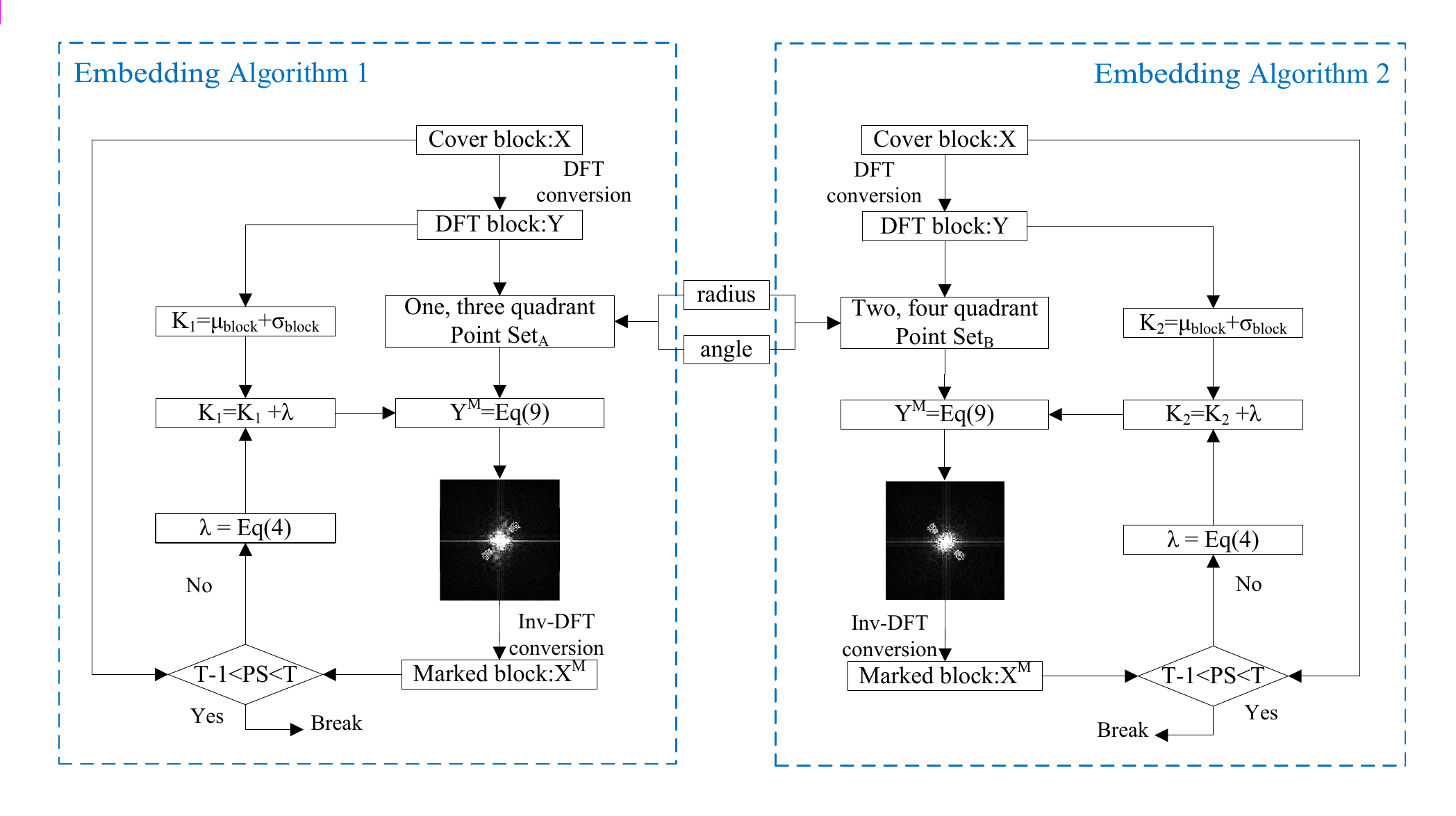}
	\caption{The procedure of embedding algorithm 1 and embedding algorithm 2.}
	\label{fig:Mark_alg}
\end{figure*}

the embedding operations will be conducted on $a_i$ and $b_j$. The embedding process is shown in Fig.~\ref{fig:Mark_alg}, and it is implemented as follows:


Step 1, selecting the embedding location. For blocks in a special group, the embedding locations are same with each other, and then, one location map for watermarking in each group is enough. We give the embedding radius $\rho$ and angle $\theta$, $\rho \in (r_1,r_2) $ and $\theta \in (\varphi_1,\varphi_2 ) \cup (\varphi_1 + \pi,\varphi_2 + \pi )$, where $0<r_1<r_2<l/2$, $0<\varphi_1<\varphi_2<\pi/2$. It is noted that we add $\pi/2$ to $\theta$ in $GroupB$ and keep $\theta$ constant in $GroupA$.
\begin{equation}
\theta =\theta +\pi /2,\ X\in GroupB
\label{eq:angle}
\end{equation}
where $X$ denoted a sub-block.
We denote the embedding locations in each block belonging to $GroupA$ as $set_A$, and
the embedding locations in each blocks belonging to $GroupB$ as $set_B$, which is denoted as  $Set = Set_A \cup Set_B$, where $Set_A=\left\{ \left( r_1,\varphi _1 \right) ,\left( r_2,\varphi _2 \right) ,...,\left( r_n,\varphi _n \right) \right\} $, where $n>0$ is number of embedding location. 


Step 2, updating embedding intensity. We modify magnitude of DFT coefficients in $Set_A$ or $Set_B$ to embed watermarks. The embedding intensity $K_1$ and $K_2$ are calculated like \crefrange{eq:fft}{eq:K1K2}. 
\begin{equation}
Y=FFT\left( X \right) 
\label{eq:fft}
\end{equation}
where $FFT$ represents the fast Fourier transform.
\begin{equation}
K_1,K_2=\mu \left( |Y| \right) +\sigma \left( |Y| \right) 
\label{eq:K1K2}
\end{equation}
where the $\mu$, $\sigma$ and $||$ denote the function to calculate the mean, standard deviation and modulo operation respectively. 

Generally, bigger embedding intensity results to low image transparency, so we introduce factor $\lambda$ to balance the embedding intensity and image transparency, and the initial value of $\lambda$ is 0. The factor $\lambda$ can be generated:
\begin{equation}
\lambda =\left\{ \begin{array}{l}
	-\sigma \left( |Y| \right) /15,\,\,if\,\,PS>T\\
	\sigma \left( |Y| \right) /15,\ \ if\,\,PS<-1\\
\end{array} \right. 
\label{eq:lamuda}
\end{equation}
where $PS$ denote the PSNR value of marked block, $T$ is a predefined threshold. It follows \crefrange{eq:PS}{eq:mse}
\begin{equation}
PS=PSNR\left( X^M,X \right) 
\label{eq:PS}
\end{equation}
where $X^M$ represents the marked block.
\begin{equation} 
PSNR=10\times \log _{10}\frac{255\times 255}{MSE}
\label{eq:PSNR}
\end{equation}
\begin{equation}
MSE=\frac{1}{b\times h}\sum_{i=1}^b{\sum_{j=1}^h{\left( c_{ij}-w_{ij} \right)^2}}
\label{eq:mse}
\end{equation}
where $c_{ij}$ and $w_{ij}$ are the cover pixels and marked pixels at $i$th and $j$th positions, respectively.

$K_1$, $K_2$ is updated as \cref{eq:updataK}:
\begin{equation}
\left\{ \begin{array}{l}
	K_1=K_1+\lambda\\
	K_2=K_2+\lambda\\
\end{array} \right. 
\label{eq:updataK}
\end{equation}

Step 3, the watermarks are embedded as \crefrange{eq:eb}{eq:ifft}.
\begin{equation}
Y^M=\left\{ \begin{aligned}
	K_1,\,\,Y_{(Set_A)}\\
	K_2,\,\,Y_{(Set_B)}\\
\end{aligned} \right. 
 \label{eq:eb}
 \end{equation}
 where $Y_{(Set_A)}$ represents the watermarked locations in $Set_A=\left\{ \left( r_1,\varphi _1 \right) ,\left( r_2,\varphi _2 \right) ,...,\left( r_n,\varphi _n \right) \right\} $.
 \begin{equation}
 X^M=IFFT\left( Y^M \right) 
 \label{eq:ifft}
 \end{equation}
 where the $IFFT$ is the corresponding inverse transform for $FFT$. 

Step 4, if the $PS\in \left[ T-1,T \right]$. Turn to next sub-blocks $a_i$ or $b_j$, otherwise goto step 2. When the embedding is finished, the sub-blocks $X$ are substituted by the marked sub-blocks $X^M$.


\subsection{Automatic watermark localization}
\begin{figure}[!htbp]
	\centering
	\includegraphics[width=3 in]{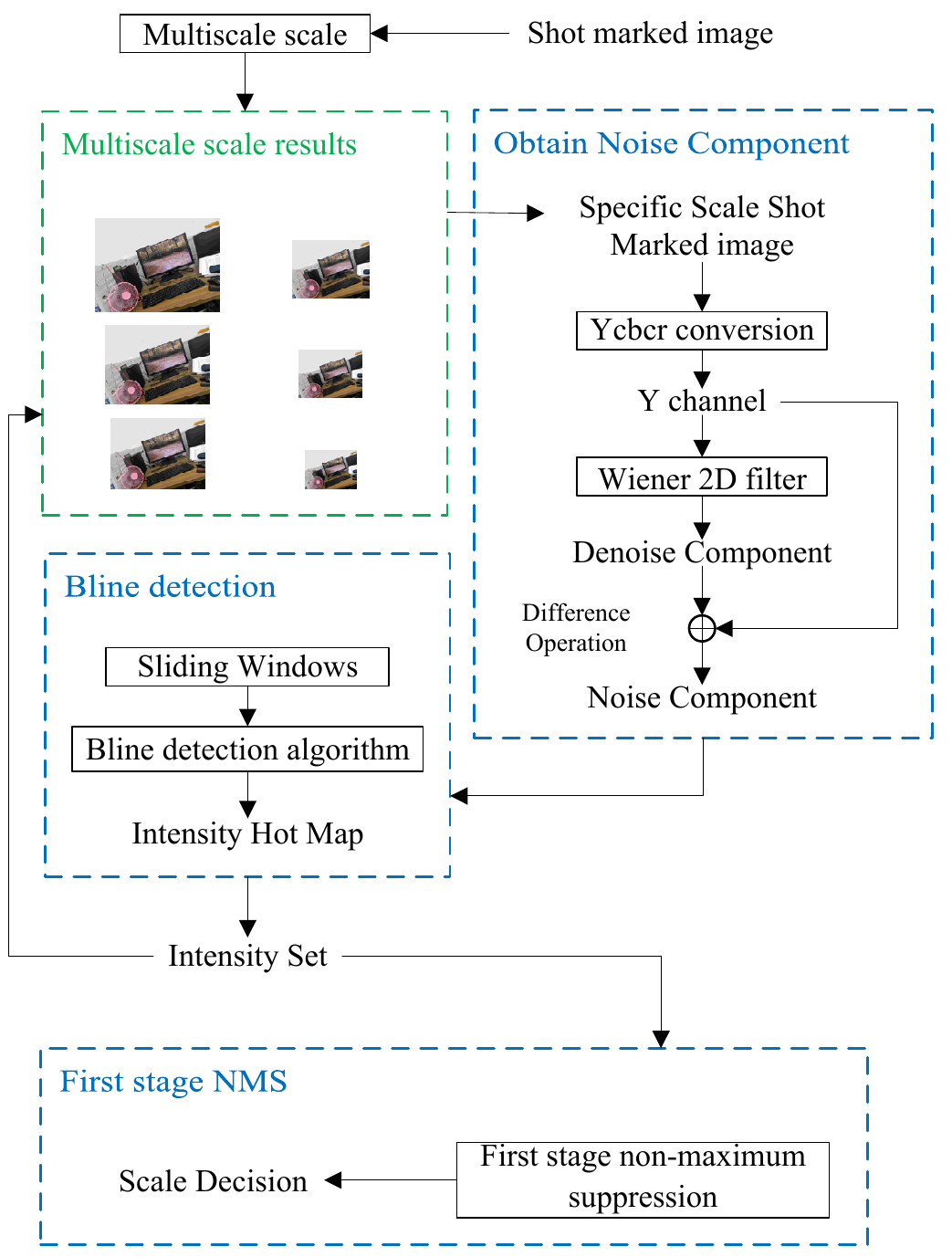}
	\caption{The framework of proposed automatic watermark localization.}
	\label{fig:BDA}
\end{figure}

The scheme that we proposed is capable of detecting the embedded watermarks in a SSI without any prior knowledge, while being robust to various attacks, including RST, perspective transform distortion, illumination changed distortion, and moir´e pattern distortion. To ensure that the watermarking can be retrieved accurately, we need to conduct some pre-processing steps, which involve identifying the suitable scale. The framework of our automatic watermark localization is depicted in Fig.~\ref{fig:BDA}.

\subsubsection{Multi-scale scale and obtain noise component}
The shot image $S$ will conduct multi-scale scaling with a scale range of 0.2 to 1 with a 0.1 interval, it can be expressed: $s_i=Scaling\left( S \right) ,i\in \left\{ 1,2\cdots ,9 \right\}$.  Then, we convert $s_i$ in a particular scale to Y-channel as the host image. The watermarking signal can be regarded as noise \cite{voloshynovskiy2000generalized}, so a wiener 2D filter will be used in the host image to get the noise component. The noise component $N$ is calculated as: 
\begin{equation}
N=s_i-wiener\left( s_i \right) 
\label{eq:wiener2d}
\end{equation}

\subsubsection{Blind detection}
 Sliding windows with fixed sizes on $N$ to detect watermarking signal intensity. Meanwhile, a proposed blind detection algorithm will conduct in each windows block. The procedure can be summarized as follows: 

Step 1: For a specific window block, we convert it to the DFT domain and obtain its magnitude coefficients :
\begin{equation}
N^m=|FFT\left( N \right) |
\end{equation}
where $N^m$ denote the magnitude coefficients in DFT domain.

Step 2: identifying watermark signal location, and save them into
$Set_A'$ and $Set_B'$. $Set_A'$ and $Set_B'$ are calculated
similarly as $Set_A, Set_B$ based on initial $\rho$ and $\theta$. But in detect process,  the group category is unknown, so we calculate the $Set_A'$ and $Set_B'$ in every window block respectively. It is noted that we expanded the angle $\theta$ to fit for the geomorphing caused by  perspective transform distortion. We refer the expanded angle $\theta$ as $\theta_a \in (\varphi_1 -\psi ,\varphi_2 + \psi)$, and $\theta_b \in (\varphi_1 -\psi+\pi/2 ,\varphi_2 + \psi +\pi /2)$ respectively, where $\psi$ is set to $5\times \pi /180$. 

Step 3: collecting a maximum of 45 magnitude coefficients s.t. $Set_A'$ and $Set_B'$ respectively. It can be expressed:
\begin{equation}
\left\{ \begin{array}{l}
	Max_1=\bigcup\limits_1^{45}{max\ N_{\left( Set'_A \right)}^{m}}\,\\
	Max_2=\bigcup\limits_1^{45}{max\ N_{\left( Set'_B \right)}^{m}}\\
\end{array} \right. 
\label{eq:MAB}
\end{equation}
where $\bigcup\limits_1^{45}max$ denotes fetching the top 45 max magnitude coefficients value.

Step 4: we use the mean value of $Max_1$ and $Max_2$ as a threshold to generating the intensity hot map (IHM) which can help to remain what approximate scale:
\begin{equation}
mPivot=\frac{\sum_{i=1}^{45}{Max_{1,i}+Max_{2,i}}}{90}
\label{eq:mPivot}
\end{equation}

Step 5: denoting numbers of elements $N_1$, $N_2$ as the number of elements that bigger than thresholds, and calculating the summation of them as $C_1$, $C_2$:
\begin{equation}
\left\{ \begin{array}{l}
	N_1=\sum{Max_1\ge mPivot}\\
	N_2=\sum{Max_2\ge mPivot}\\
\end{array} \right. 
\label{eq:N1N2}
\end{equation}

\begin{equation}
\left\{ \begin{array}{l}
	C_1=\sum_{i=1}^{45}{Max_{1,i}\ge mPivot}\\
	C_2=\sum_{i=1}^{45}{Max_{2,i}\ge mPivot}\\
\end{array} \right. 
\label{eq:C1C2}
\end{equation}

Step 6: we use the difference between $N_1$ and $N_2$ to mark watermarks.
\begin{equation}
d=N_1-N_2
\label{eq:DIFF}
\end{equation}

Step 7: IHM will be generated based on $d$,$mPivot$,$C_1$ and $C_2$, which denotes intensity of the watermarking signal. It can be expressed:
\begin{equation}
IHM=\left\{ \begin{array}{l}
	\begin{aligned}
	&0,if\ abs\left( d \right) <T\\
	&sum\left(C_1 \right) \times d,if\ T<abs\left( d \right) \& d<0\\
	&sum\left(C_2 \right) \times d,if\ T<abs\left( d \right) \& d>0\\
	\end{aligned}
\end{array} \right. 
\label{eq:IHM}
\end{equation}
where $T$ denotes a empirical Threshold. In this experiments, the $T$ is initialized as 4.

\begin{figure*}[!htbp]
	\centering
	\subfigure[]{\includegraphics[height=1.4 in]{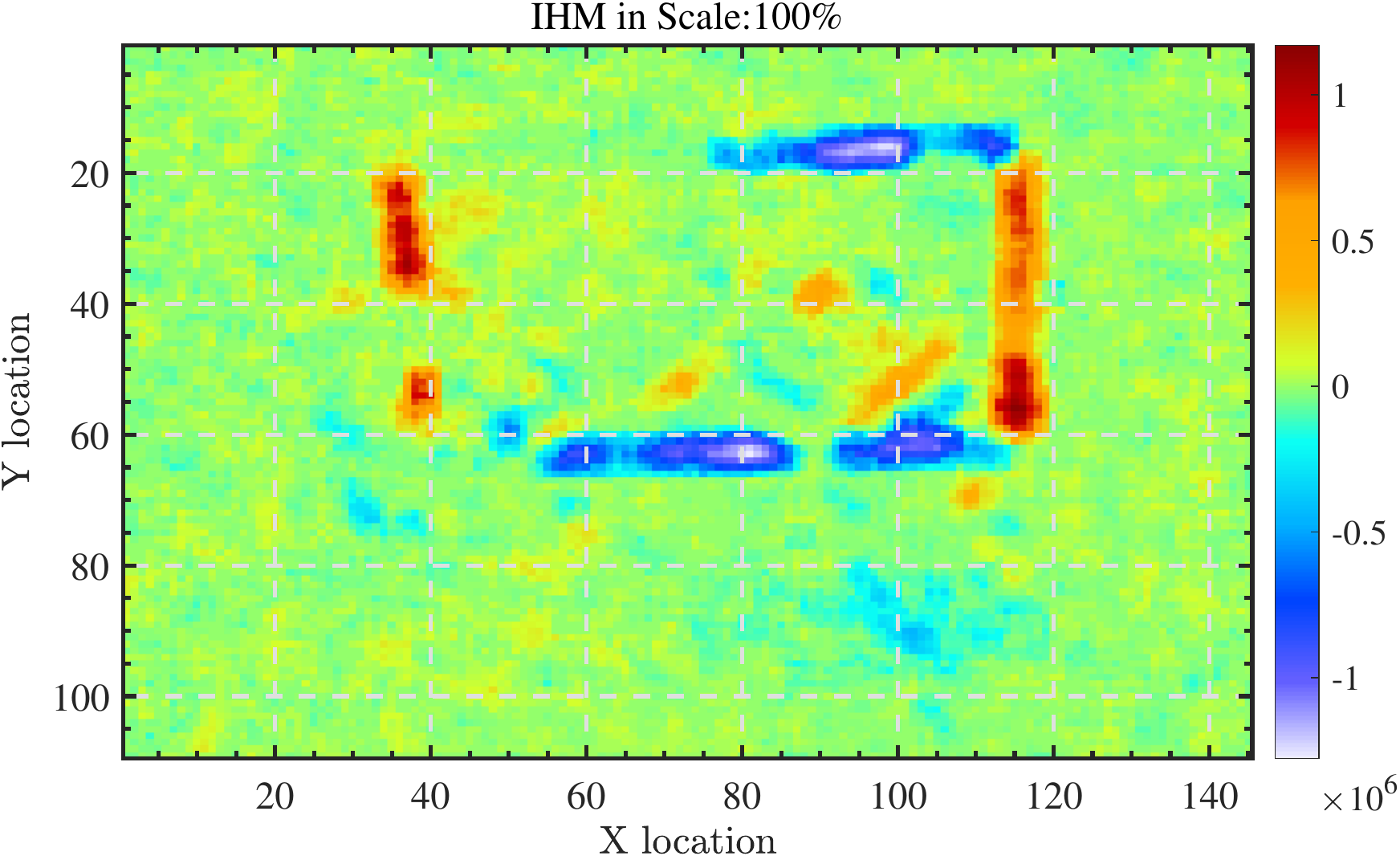}}
	\hspace{-6 pt}
	\vspace{-5 pt}
	\subfigure[]{\includegraphics[height=1.4 in]{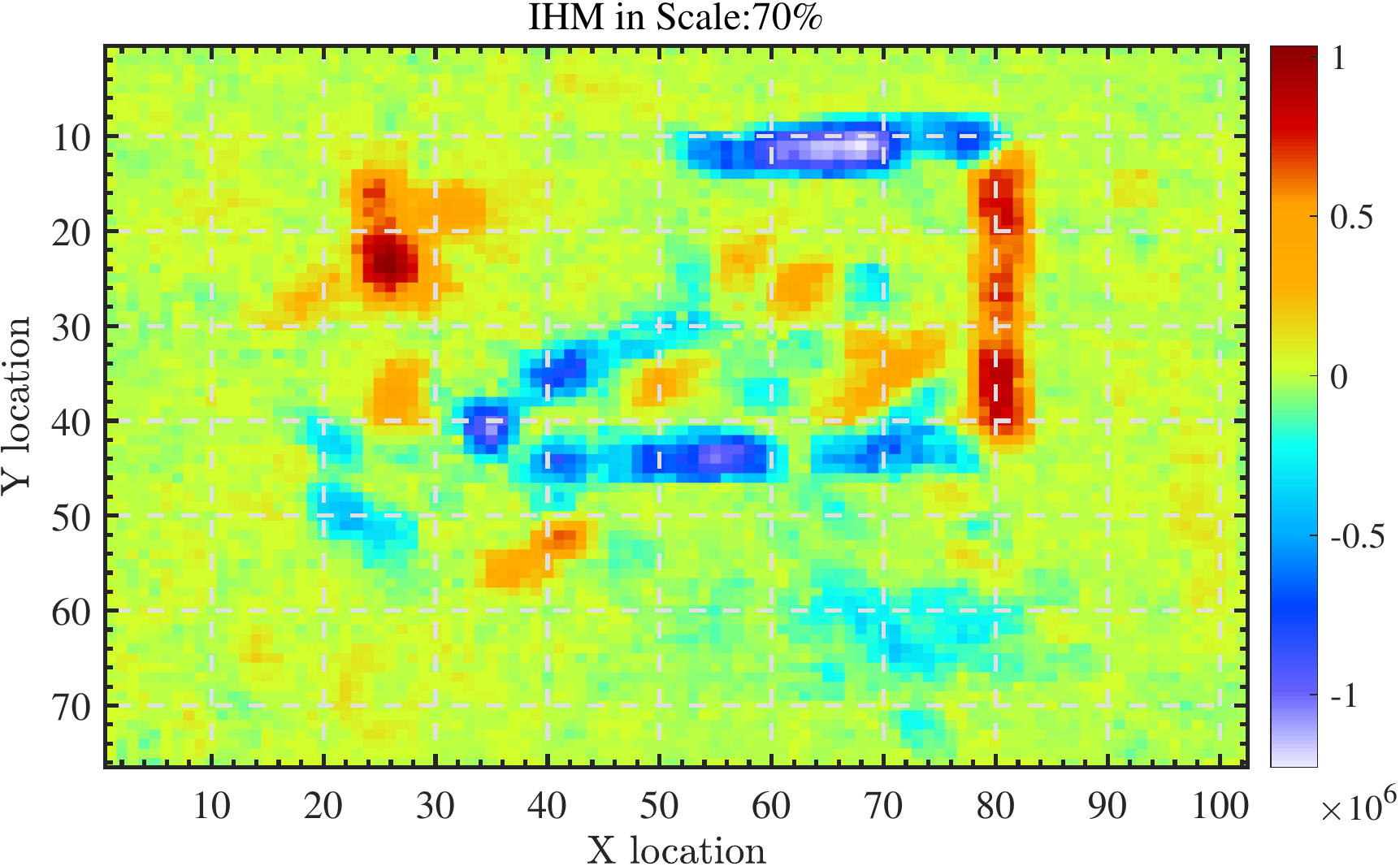}}
	\hspace{-6 pt}
	\vspace{-5 pt}
	\subfigure[]{\includegraphics[height=1.4 in]{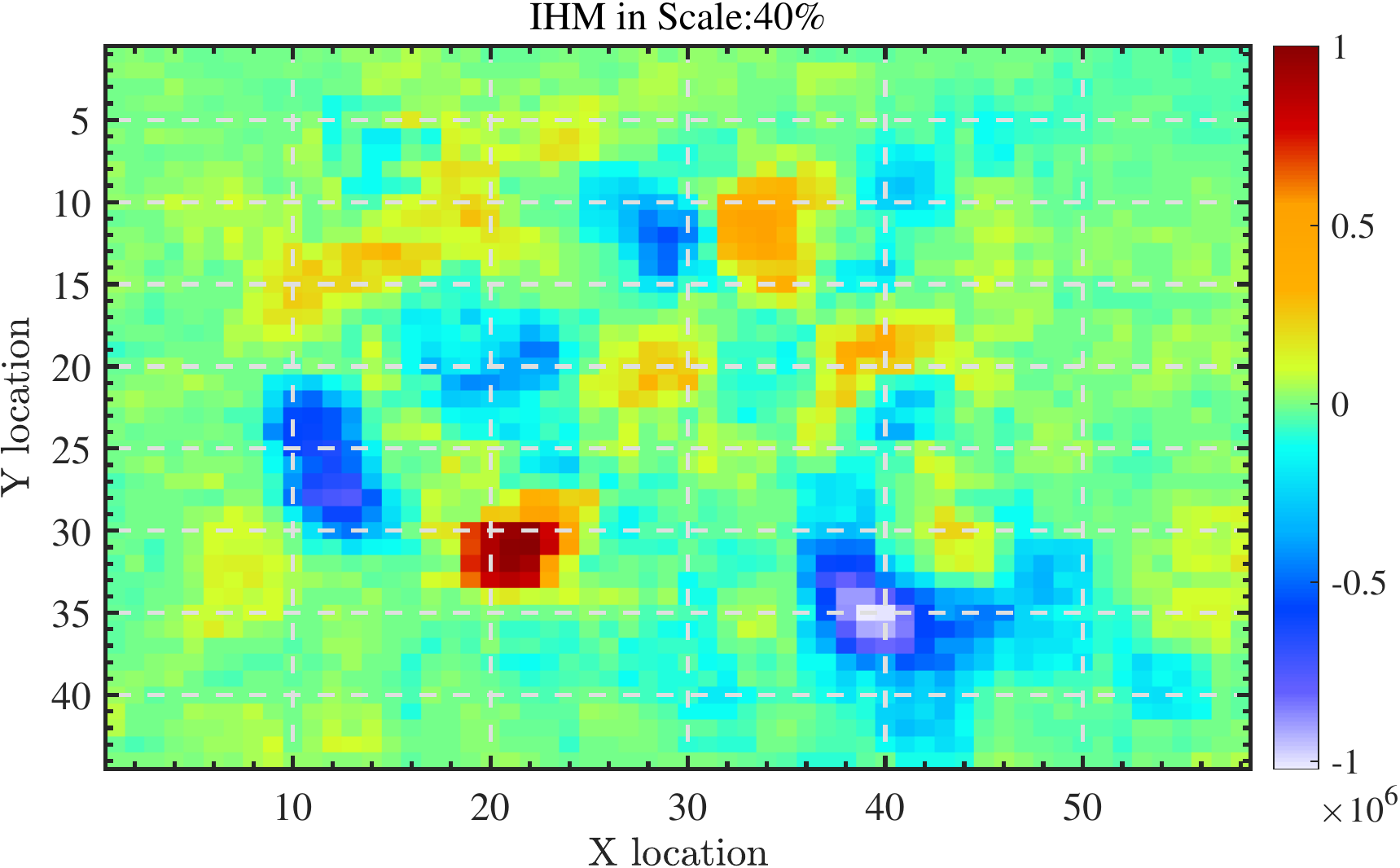}}
	\hspace{-6 pt}
	\vspace{-5 pt}
	\\
	\subfigure[]{\includegraphics[height=1.4 in]{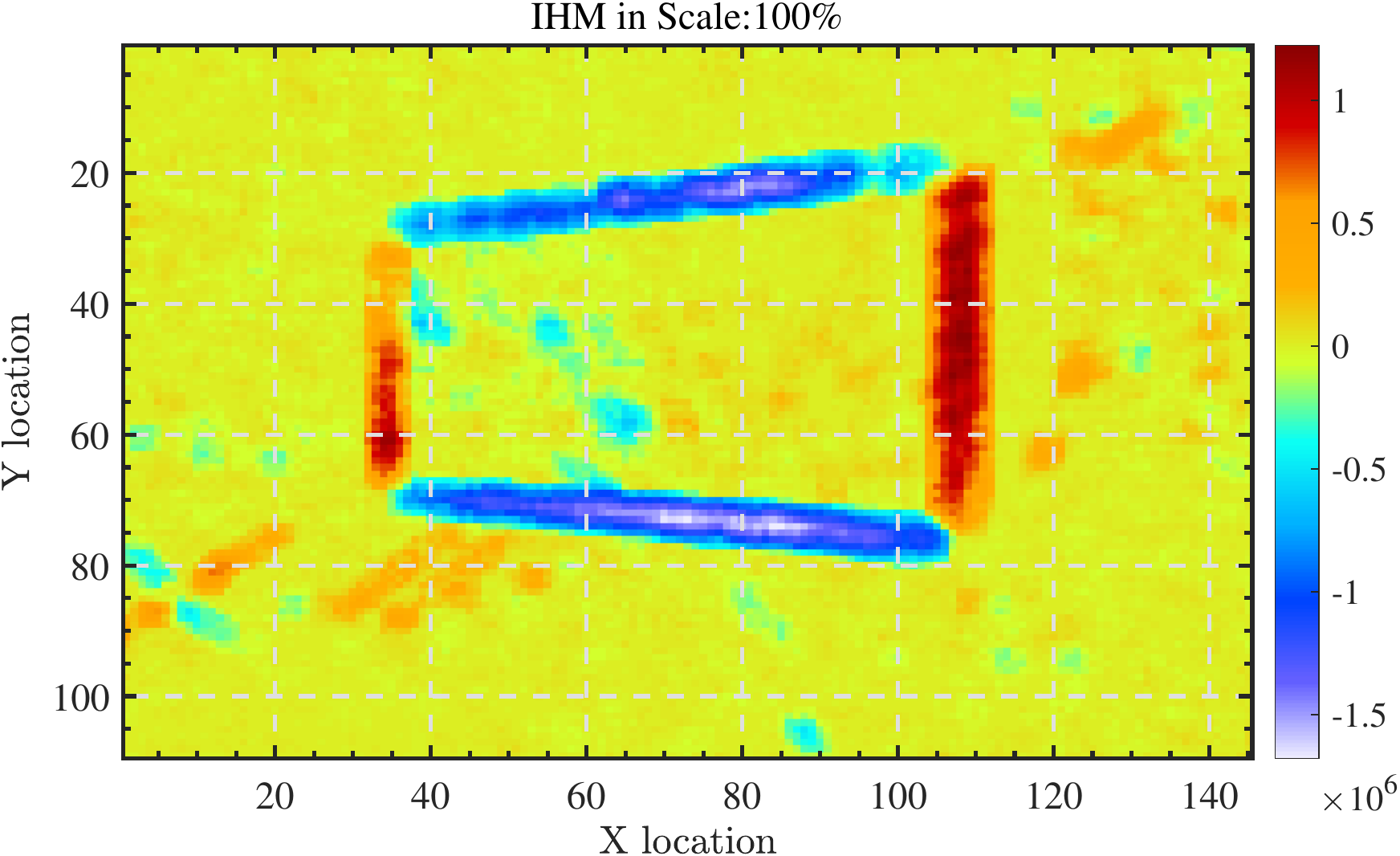}}
	\hspace{-6 pt}
	\vspace{-5 pt}
	\subfigure[]{\includegraphics[height=1.4 in]{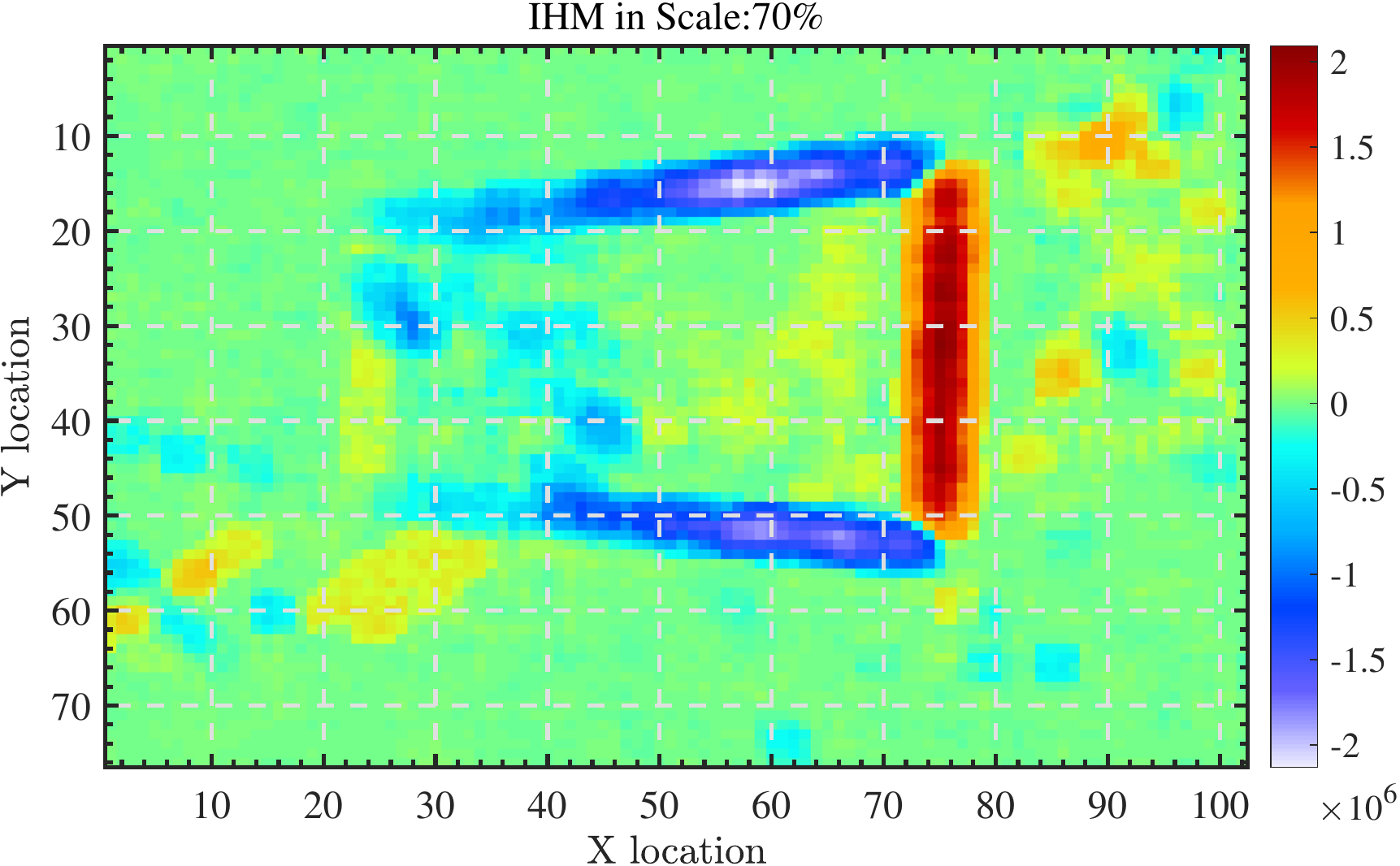}}
	\hspace{-6 pt}
	\vspace{-5 pt}
	\subfigure[]{\includegraphics[height=1.4 in]{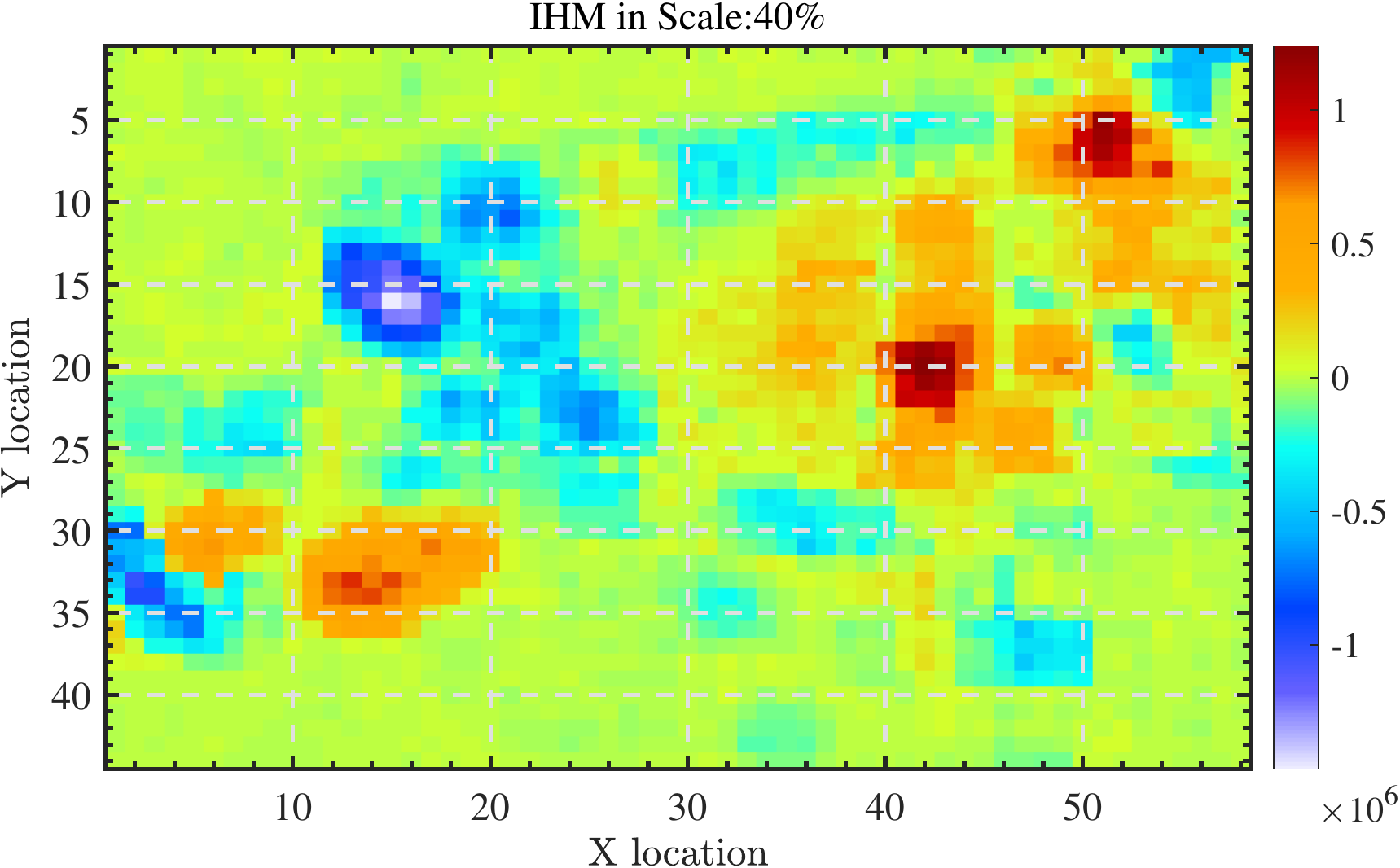}}
	\hspace{-6 pt}
	\vspace{-5 pt}
	\\
	\subfigure[]{\includegraphics[height=1.4 in]{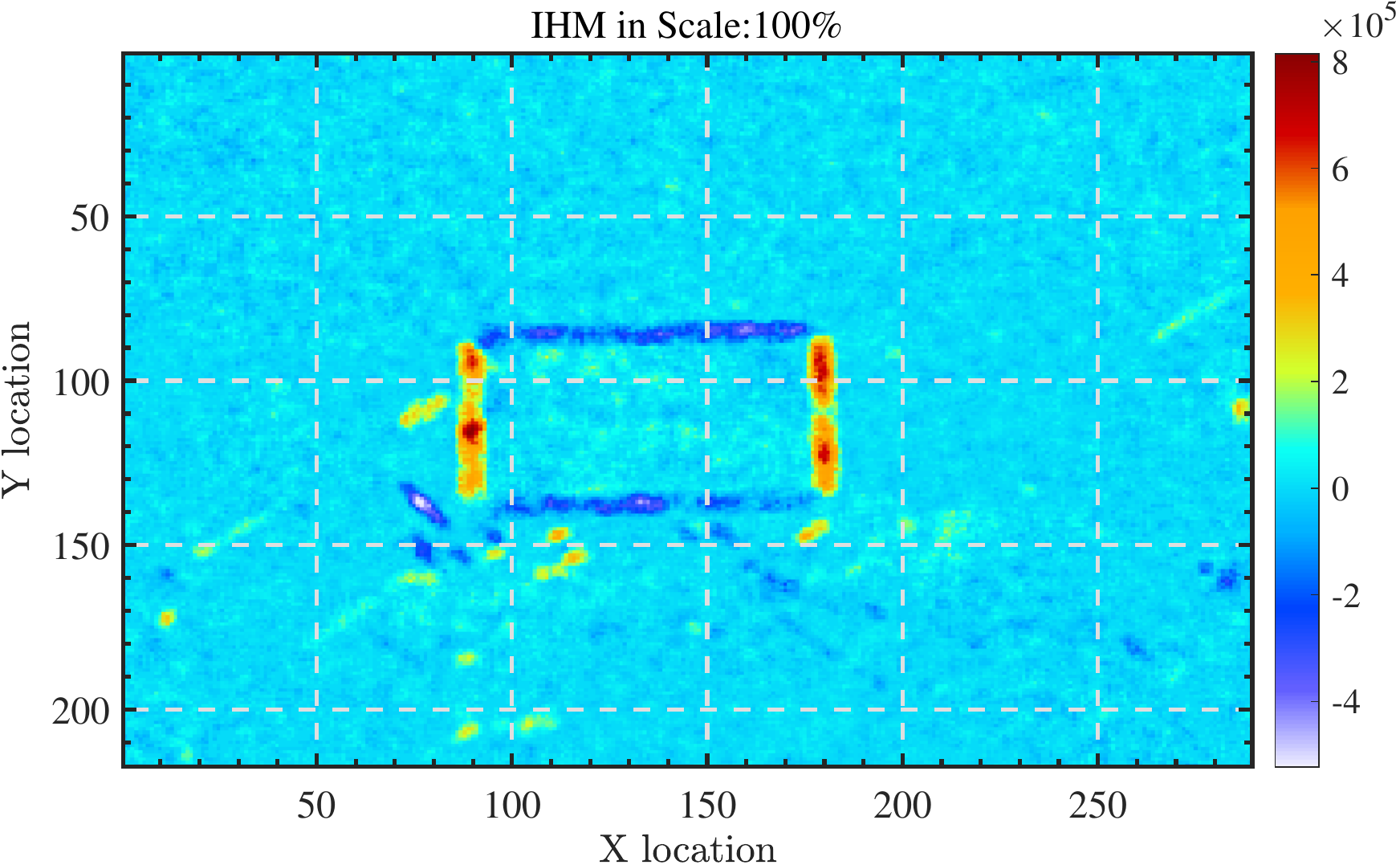}}
	\hspace{-6 pt}
	\vspace{-5 pt}
	\subfigure[]{\includegraphics[height=1.4 in]{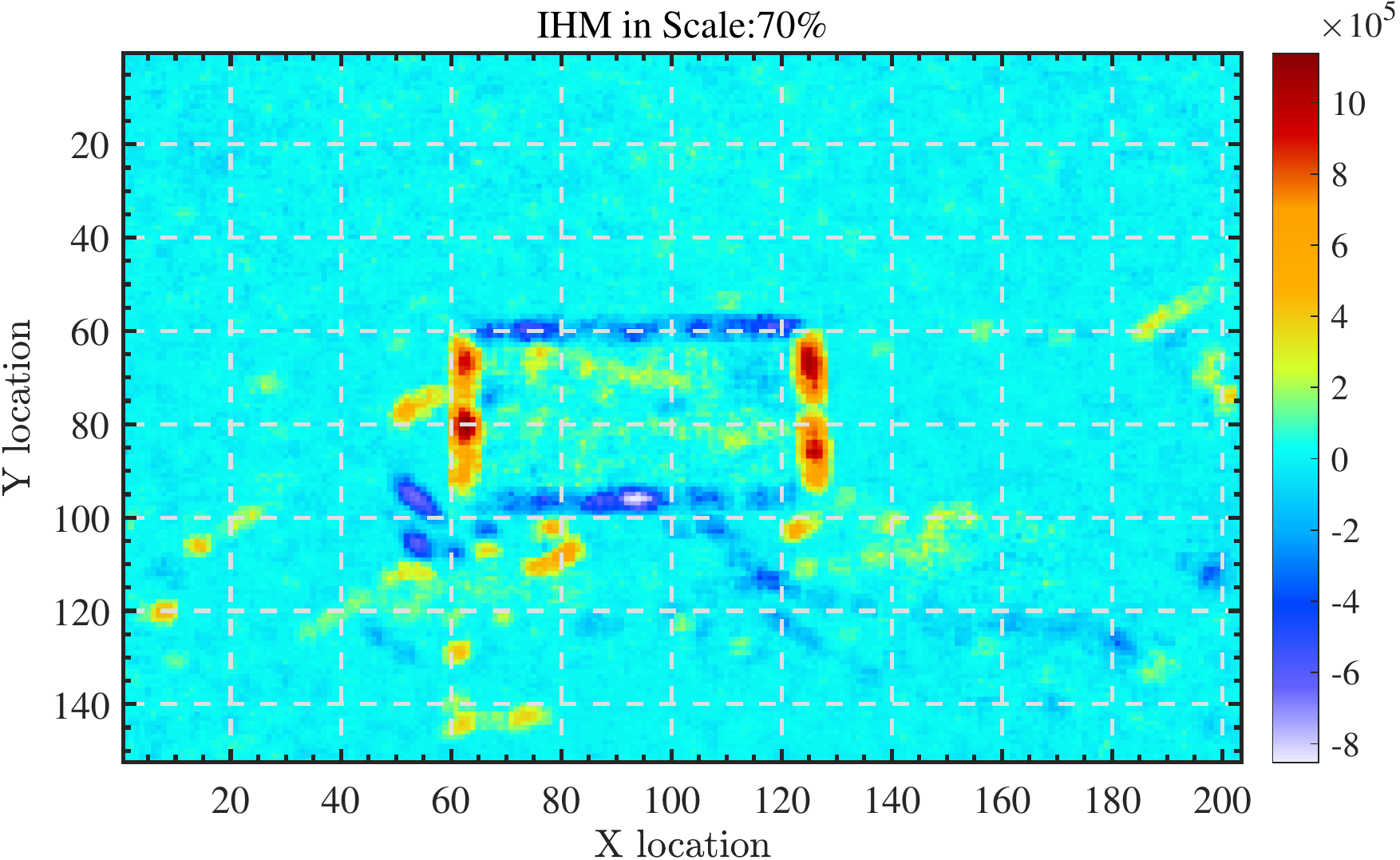}}
	\hspace{-6 pt}
	\vspace{-5 pt}
	\subfigure[]{\includegraphics[height=1.4 in]{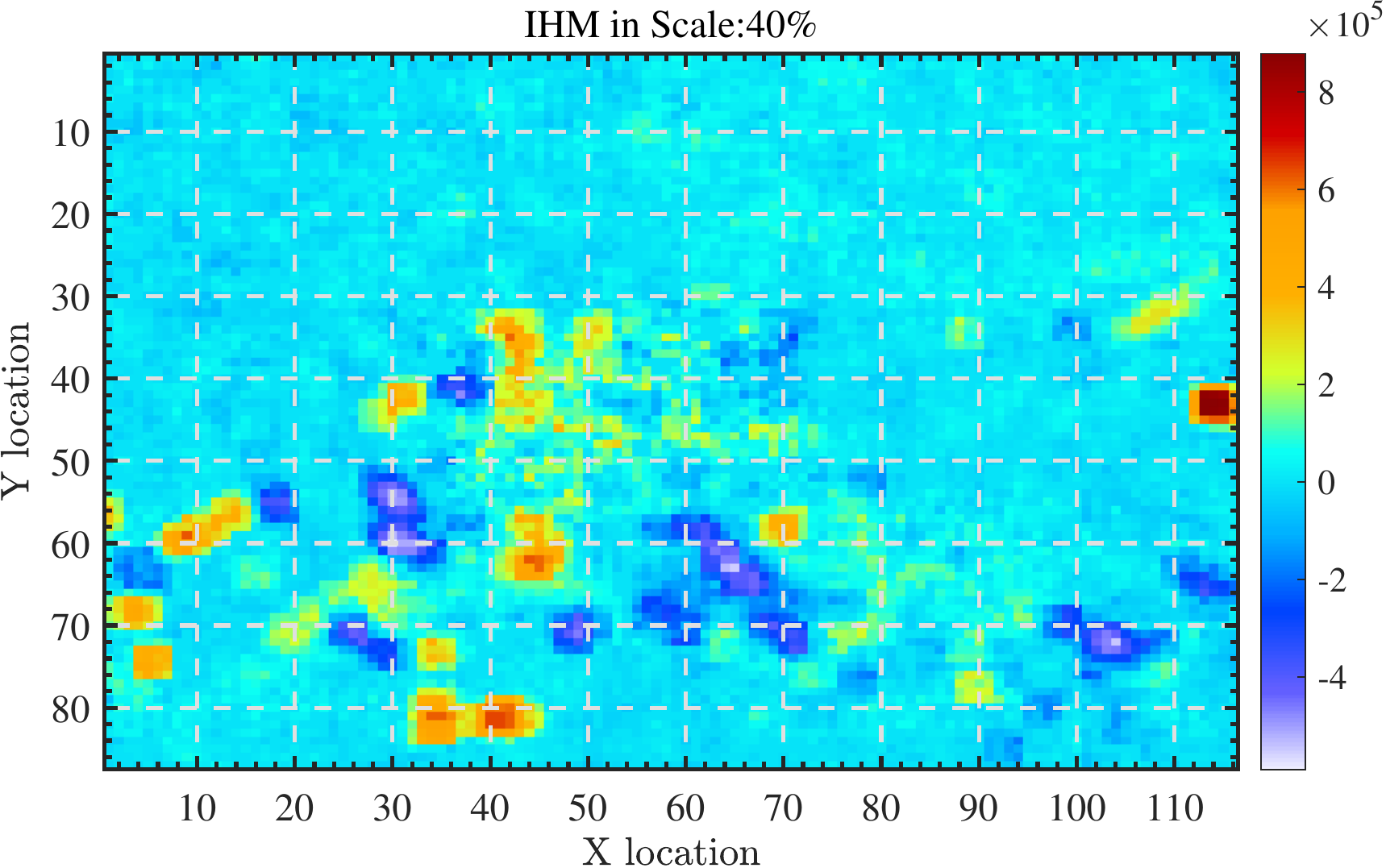}}
	\hspace{-6 pt}
	\vspace{-5 pt}
	\caption{The visual IHM results in some sample, where each row denote one SSI and each col denote visual IHM.}
	\label{fig:IHMV}
\end{figure*}

The part of visual IHM results can be seen in Fig.~\ref{fig:IHMV}. In some scales, we can see the shadow bounding boxes enclosed 
with red and blue, which are the private image margin. However, on an inappropriate scale, the bounding will miss.

Finally, we save them in an intensity set. When finish the multi-scale stage, the intensity set based on the blind detection algorithm will be generated, which will guide the first stage NMS process. 

\subsubsection{First stage NMS}
The first stage NMS will be applied in intensity set to obtain the most suitable scale and decide which scale should be remained for a specific SSI.

Since the actual SSI scale is unknown, so we attempt to scale SSI to get the approximate scale based on scale-intensity map (SIM) and it can be calculated by \crefrange{eq:IS}{eq:SI}.
\begin{equation}
IS=IHM_{scale},scale\in \left\{ 1,0.9,\cdots ,0.2 \right\} 
\label{eq:IS}
\end{equation}
where IS denotes Intensity Set in all scale of attempt.
\begin{equation}
SIM=\sigma \left( IS_i \right) ,i\in \left\{ 1,2\cdots ,9 \right\} 
\label{eq:SI}
\end{equation}

For screen-shooting operations, even though the same digital image is shot by different record devices at the same distance, the SSI scale is different. An example is flexible camera pixels, the SSI scale is quite different between 6400 million pixels conditions and 1200 million pixels conditions, but both of them will integrate with the most popular mobile phone.

A NMS similar idea is proposed to solve the actual SSI scale unknown issue. The local peaks in SIM will help to decide a suitable scale, and we refer to the $findpeaks$ algorithm \footnote{\url{https://ww2.mathworks.cn/help/signal/ref/findpeaks.html?s_tid=srchtitle_findpeak_1}}:
\begin{equation}
Peak_j=findpeaks\left( SIM \right) ,1\le j\le 9
\label{eq:fp}
\end{equation}

There maybe multiple $Peak$ values will be gotten using \cref{eq:fp}. But the first $Peak$'s argument is saved as scale decision. It can be expressed: 
\begin{equation}
Scale\ Decision=arg\left( Peak_{j=1} \right) 
\label{eq:DS}
\end{equation}
\subsection{Bounding box generation}

In this section, we are aiming to harness the power of max/min peaks to accomplish the regression of bounding box margin line. The crux of this idea is that even if some peaks points or watermarking regions are not sustained under screen-shooting distortion, we can still locate the region of interest (RoI) in the SSI using the surviving peaks points or watermarking regions.

\begin{figure*}[!htbp]
	\centering
	\includegraphics[width=7 in]{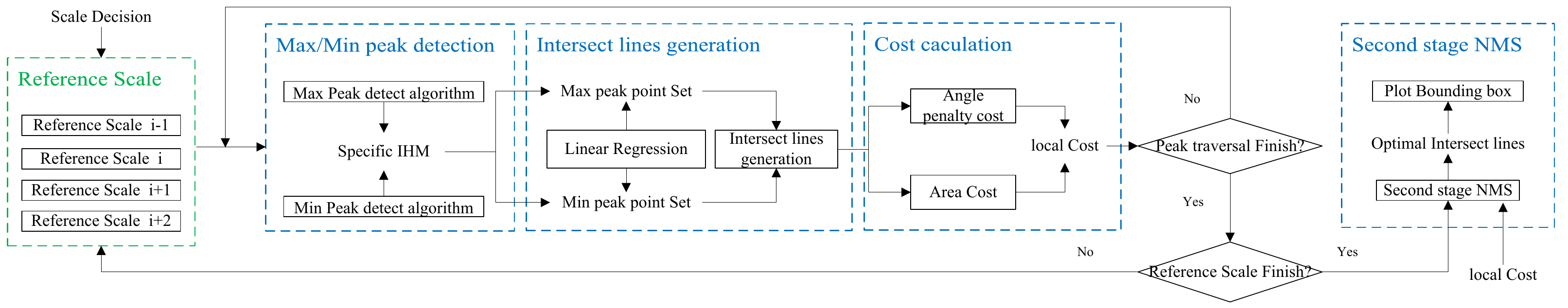}
	\caption{The framework of proposed bounding box generation.}
	\label{fig:FBB}
\end{figure*}

In this particular endeavor, we may face a predicament of being susceptible to aberrant points that may affect a particular regression undertaking. Since it is imperative to execute a linear regression fused with the traversal of the peaks, we suggest employing a second stage NMS grounded on costs which is also efficacious in counterbalancing the influence of outlying points. The framework depicting the process of bounding box generation is demonstrated in Fig.~\ref{fig:FBB}.

\subsubsection{Max/Min peak detection}
Here we select 4 IHM based on $scale\ decision$. For a specific IHM, the max peak detection algorithm and min peak detection algorithm will apply to get max and min peak point set respectively.

The max/min peaks detecting algorithm can be described as following:

Step 1: The coordinates of max/min value in IHM should be extracted using \cref{eq:MAXMIN}
\begin{equation}
\left\{ \begin{array}{l}
	(x_1,y_1)=arg\max \left( IHM \right)\\
	(x_2,y_2)=arg\min \left( IHM \right)\\
\end{array} \right. 
\label{eq:MAXMIN}
\end{equation}

Step 2: The peak neighborhood in IHM should be cleaned using \crefrange{eq:expcor}{eq:PEZ}
\begin{equation}
\begin{array}{c}
	x_a=\left\{ x_1,x_1\pm 1 \right\} ,y_a=\left\{ y_1,y_1\pm 1 \right\}\\
	x_b=\left\{ x_2,x_2\pm 1 \right\} ,y_b=\left\{ y_2,y_2\pm 1 \right\}\\
\end{array}
\label{eq:expcor}
\end{equation}
\begin{equation}
\left\{ \begin{array}{l}
	IHM_{x_a,y_a}=0\\
	IHM_{x_b,y_b}=0
\end{array} \right. 
\label{eq:PEZ}
\end{equation}

Step 3: In general, we need to repeatedly conduct steps 1 and step 2 to get sufficient $(x_1,y_1)$ and $(x_2,y_2)$ and an empirical threshold $
\in \left[ \alpha ,\beta \right]$ to output required points. In the experiment, the $\alpha = 13$ , $\beta = 18$. The collection points in each round is saved in max peak point set and min peak point set respectively. The procedure can be expressed:
\begin{equation}
\left\{ \begin{array}{l}
	Max_{Peak\ Set}=x_{1,i},y_{1,i}\\
	Min_{Peak\ Set}=x_{2,j},y_{2,j}\\
\end{array}i,j\in \left[ \alpha ,\beta \right] \right. \cap N
\label{eq:label}
\end{equation}
where $N$ denote the natural number set.
\subsubsection{Intersect Lines Generation}
\begin{figure}[!htbp]
	\centering
	\subfigure[a]{\includegraphics[height=1.0 in]{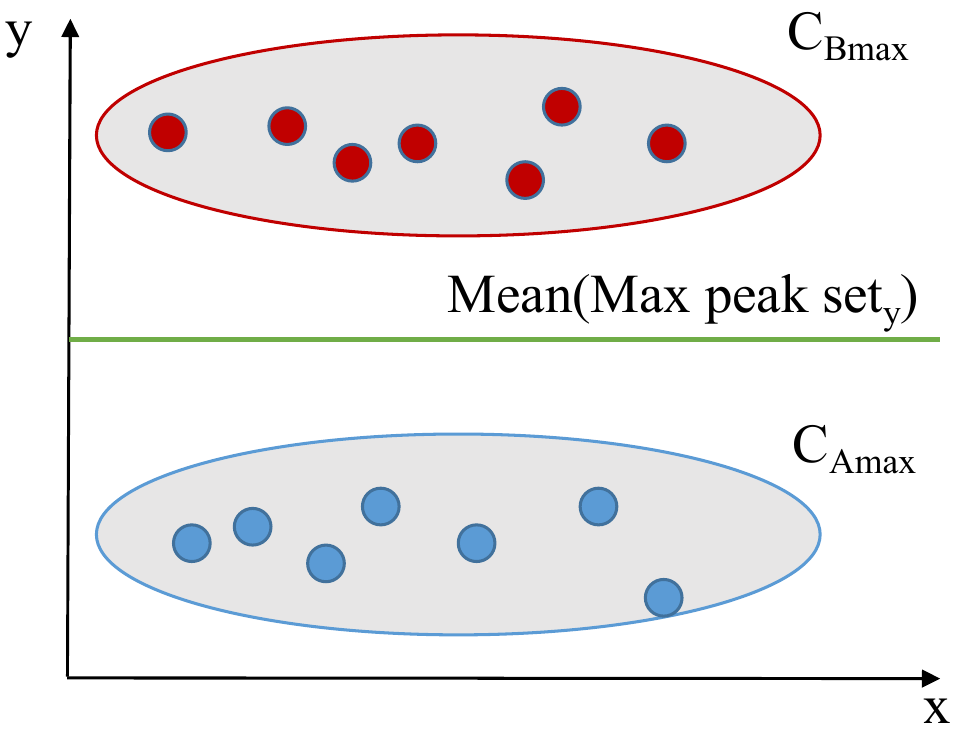}}
	\hspace{0 pt}
	\subfigure[b]{\includegraphics[height=1.0 in]{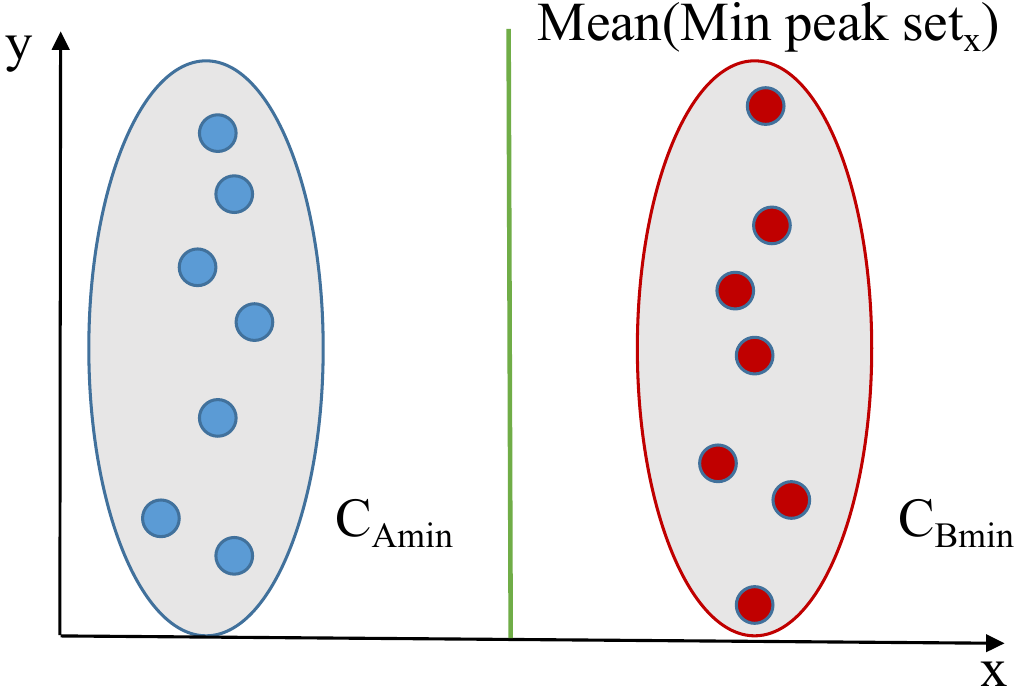}}
	\caption{A distance-based clustering example. (a) The results using \cref{eq:MaxDclus} . (b) The results using \cref{eq:MinDclus} .}
	\label{fig:CAB}
\end{figure}

The $Max_{Peak\ Set}$ and $Min_{Peak\ Set}$ are divided based on distance clustering before line regression to generate intersect lines. The procedure can be expressed as follows:

Step 1: distance-based clustering will be conducted in $Max\ peak\ set$ and $Min\ peak\ set$ respectively, the methods can be expressed:
\begin{equation}
\left\{ \begin{array}{l}
	C_{AMax}=y_{1,i}<mean\left( Max\ peak\ set_y \right)\\
	C_{BMax}=y_{1,i}>mean\left( Max\ peak\ set_y \right)\\
\end{array} \right. 
\label{eq:MaxDclus}
\end{equation}
\begin{equation}
\left\{ \begin{array}{l}
	C_{AMin}=x_{2,j}<mean\left( Min\ peak\ set_x \right)\\
	C_{BMin}=x_{2,j}>mean\left( Min\ peak\ set_x \right)\\
\end{array} \right. 
\label{eq:MinDclus}
\end{equation}
A example for distance-based clustering is shown in Fig.~\ref{fig:CAB}.

Step 2:A simple linear regression will be applied in $C_{AMax},C_{BMax},C_{AMin},C_{BMin}$ respectively to generate intersect lines. It can be expressed:
\begin{equation}
\left\{ \begin{array}{l}
	\ell _a=lr\left( C_{AM\text{ax}} \right)\\
	\ell _b=lr\left( C_{BMax} \right)\\
	\ell _c=lr\left( C_{AMin} \right)\\
	\ell _d=lr\left( C_{DMin} \right)\\
\end{array} \right. 
\label{eq:label}
\end{equation}
where the $\ell$ and $lr$ denote the a line and linear regression respectively.

\begin{figure}[!h]
 	\centering
 	\includegraphics[width=3 in]{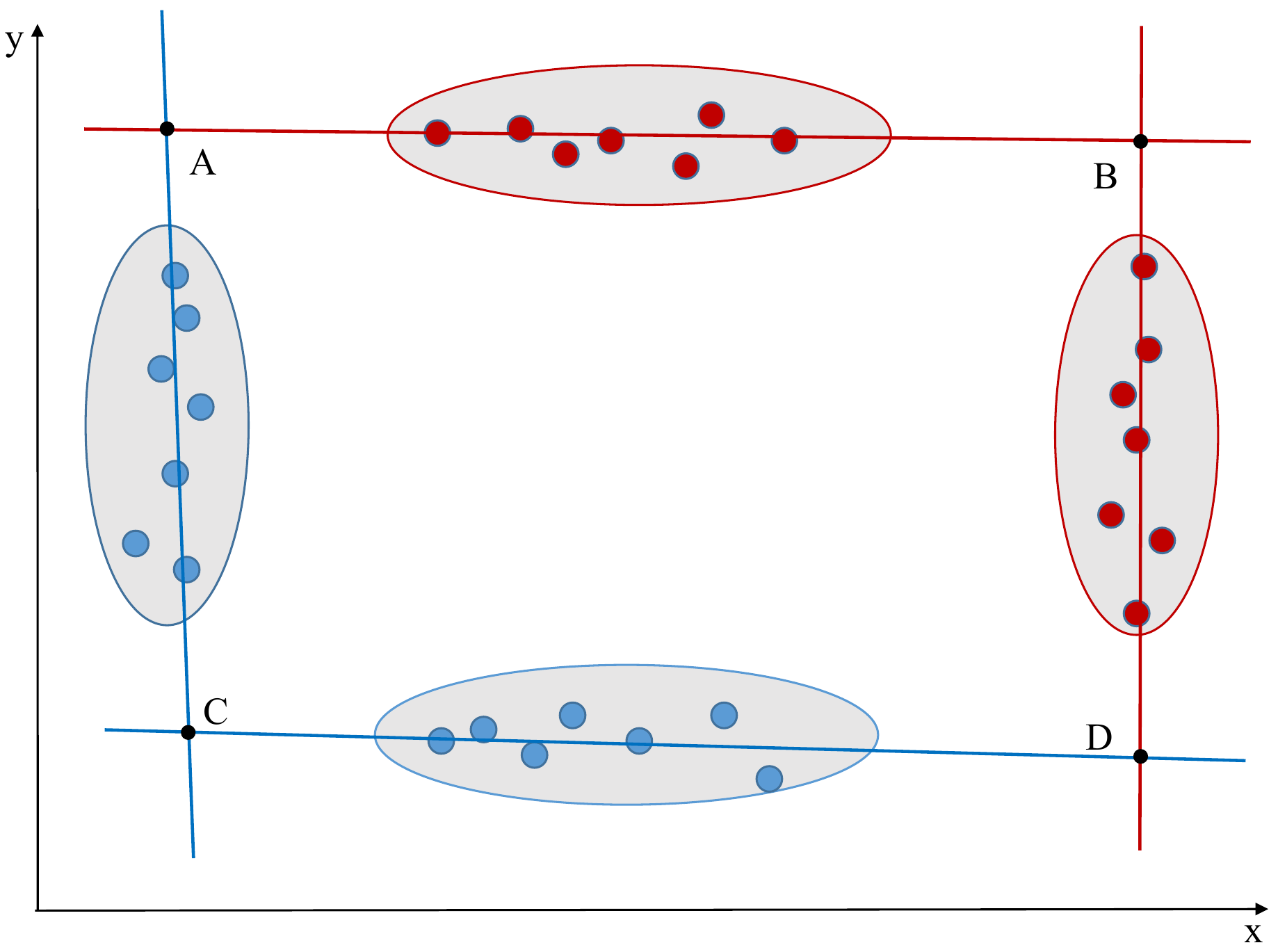}
 	\caption{A example for Intersect Lines generate, where A,B,C,D denote the corners of intersection respectively.}
 	\label{fig:LC}
 \end{figure}

Step 3: We calculate the four corners of intersection by $\ell$, and an example for the procedure is shown in Fig.~\ref{fig:LC}.

\subsubsection{Cost calculation}
The generated enclosed regions are preliminary results, but the results are not always perfect such as Fig.~\ref{fig:LC}. Therefore, we need to evaluate the accuracy of each enclosed region by introduced angle penalty cost and area cost. Angle penalty cost and area cost will be introduced respectively to generate the final local cost as \cref{eq:LC}. The local cost will guide the second stage NMS to select the optimal intersection line.

Step 1: Angle penalty cost($AP_{cost}$) is to constant the angle of the regions. For a specific SSI, the regions of the private image approximate a rectangle, on the other hand, the four corners in a quadrilateral are close to 90\degree. There is an example to calculate the angle penalty cost in Fig.~\ref{fig:LC} of corners of B. The direction vector will be calculated by using \cref{eq:e1e2}
\begin{equation}
\left\{ \begin{array}{l}
	\vec{e}_1=\overrightarrow{\ell }_{A\rightarrow B}/|\overrightarrow{\ell }_{A\rightarrow B}|\\
	\vec{e}_2=\overrightarrow{\ell }_{D\rightarrow B}/|\overrightarrow{\ell }_{D\rightarrow B}|\\
\end{array} \right. 
\label{eq:e1e2}
\end{equation}

Step 2: Vector product with $\vec{e}_1$ and $\vec{e}_2$:
\begin{equation}
VP_B=\vec{e}_1\cdot \vec{e}_2
\label{eq:Vp}
\end{equation}

Step 3: $AP_{cost}$ will be calculated with $VP_A$,$VP_B$,$VP_C$ and $VP_D$ using \crefrange{eq:VVV}{eq:Ap} 
\begin{equation}
v=\sum\limits_{i=A}^D{VP_i}
\label{eq:VVV}
\end{equation}
\begin{equation}
AP_{\cos t}=\lambda e^{-v}
\label{eq:Ap}
\end{equation}
where $\lambda$ is a factor to control $AP_{cost}$, in this experiment the $\lambda$=5.

Step 4: Area cost ($A_{cost}$) is to calculate the area of regions. It can be calculated using \crefrange{eq:SEG}{eq:AC}:
\begin{equation}
S=area\left( EG_{ABCD} \right) 
\label{eq:SEG}
\end{equation}
where $EG_{ABCD}$ and $area$ denote the enclosed regions and find its area.
\begin{equation}
A_{\cos t}=S/scale
\label{eq:AC}
\end{equation}

Step 5: the local cost(LC) is to denote the deviate degree generated enclosed region with the expected target region, and it is denoted as:
\begin{equation}
LC_{p,s}=A_{\cos t}\times AP_{\cos t}
\label{eq:LC}
\end{equation}
where $p$ and $s$ denote the index of peaks and index of reference scale respectively.
\subsubsection{Second stage NMS and plot bounding box}
The second stage NMS is to find optimal intersect line based on with the maximum local cost using \cref{eq:ol}
\begin{equation}
\ell =arg\max \left( LC_{p,s} \right) 
 \label{eq:ol}
 \end{equation}
 where $\ell$ denote the optimal intersect line.

The final detection results will use bounding box based on optimal intersect line to plot out. 

\subsection{Automatic watermarking detection}
\begin{figure}[!h]
	\centering
	\includegraphics[width=3 in]{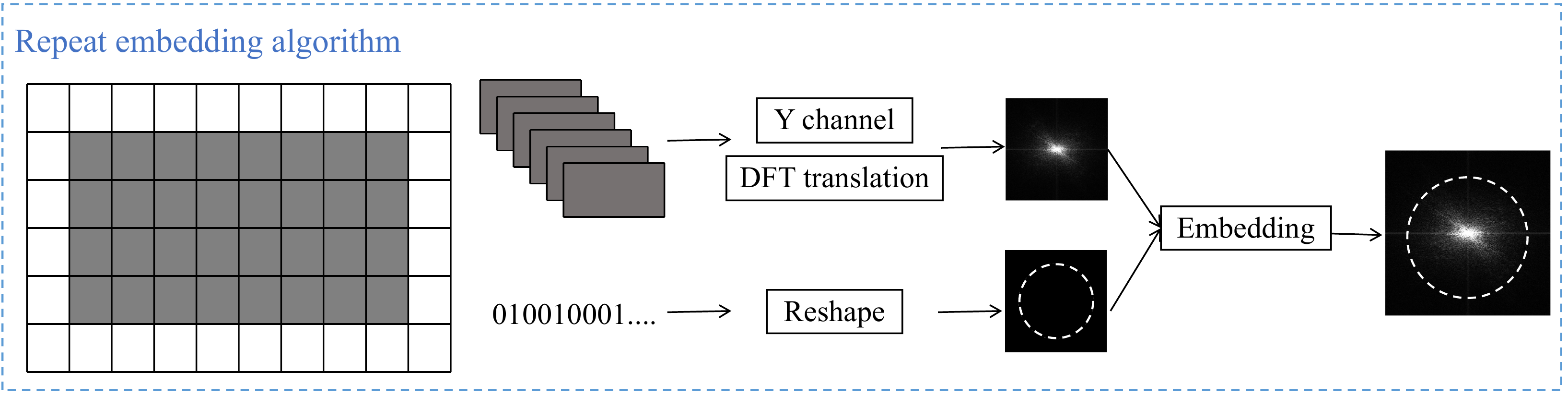}
	\caption{The pipeline of embedding framework}
	\label{fig:repeatemb}
\end{figure}

The concept in this section is to employ the automatic watermark detection (AWD) to extract the concealed watermark from the identified RoI by performing the perspective correction. Our proposed watermark embedding framework is based on the DFT to embed the forensics information in the cover. As demonstrated in Fig.~\ref{fig:repeatemb}, the pipeline of the repeat embedding framework is illustrated. The identical watermark sequence is recurrently embedded in all gray blocks. The repeat embedding algorithm is merged with the periodic extension in the DFT domain \cite{harris1978use}. This implies that we can precisely extract watermarks based on each block. Following the perspective correction process, we can successfully extract the watermark.
\subsubsection*{authentication embedding}
As shown in Fig.~\ref{fig:repeatemb}, the gray regions will be blocked to cover forensics data. For embedding algorithm we refer to \cite{kang2010efficient,poljicak2011discrete,fang2018screen}, which are keypoint-based, invariant-domain based and template-based methods respectively.
\subsubsection*{perspective correction}
\begin{figure}[!h]
 	\centering
 	\includegraphics[width=3 in]{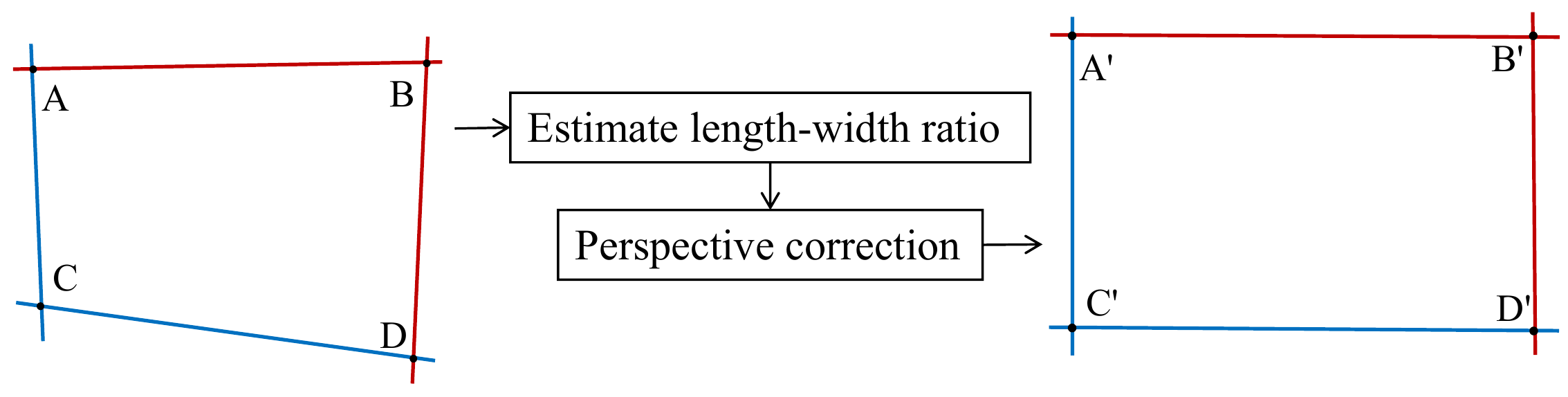}
 	\caption{A pipeline of perspective correction.}
 	\label{fig:perc}
 \end{figure}

The perspective correction based on \cref{eq:ol} is shown in Fig.~\ref{fig:perc}. The optimal intersect line will generate four corners $A(x_1,y_1)$,$B(x_2,y_2)$,$C(x_3,y_3)$,$D(x_4,y_4)$, then we need to estimate length-width ratio, it can be expressed:
\begin{equation}
\left\{ \begin{array}{l}
	width=\left( |AB|+|CD| \right) /2\\
	length=\left( |AC|+|BD| \right) /2\\
\end{array} \right. 
\label{eq:wl}
\end{equation}
where $|AB|$ denote the distance of A to B.

Then we set the transformed coordinates corresponding to these 4 corners $
A'(x'_1,y'_1),B'(x'_2,y'_2),C'(x'_3,y'_3),D'(x'_4,y'_4)$. Substituting the 8 coordinates into \cref{eq:pt} we can get 8 equations, so the value of$a_1,b_1,c_1,a_0,b_0,a_2,b_2,c_2$ can be obtained by solving these equations. 
\begin{large}
\begin{equation}
\begin{array}{l}
	x'=\frac{a_1x+b_1y+c_1}{a_0x+b_0x+1}\\
	y'=\frac{a_2x+b_2y+c_2}{a_0x+b_0x+1}\\
\end{array}
\label{eq:pt}
\end{equation}
\end{large}
With determining these parameters, we can form a mapping from the distorted image to the corrected image.
\subsubsection*{watermark extraction}
The corrected image is used to extract watermark. We overlap blocking the corrected image and extract the watermark in each block, with the window size is $length$, the stripe is 
128. The extraction is inverse precess for embedding algorithm, and its details we refer as \cite{kang2010efficient,poljicak2011discrete,fang2018screen}.
\section{Experiment}
In this section, the experimental setting will be introduced first, then extensive experiments for detection accuracy will be declared. Watermark robustness experiments are given at last. We also analyze the experimental results and introspect the limitations in our scheme, and they are discussed at last. 

\subsection{Experimental setting}
\subsubsection{Parameter Configuration}
The experimental image size is $2560 \times 1536$ and it can be download from website\footnote{Image dataset: \url{https://wallhaven.cc/}}.  In the marked stage, the block size is configured as $128 \times 128$. The embedding radius and angles are configured as 15 to 25 with an interval of 1.25 and 35 to 55 with an interval of 5 respectively. In the detection procedure stage, the sliding window size is the same as the block size. 
\subsubsection{Hardware Settings}
The monitor are $ThinkVision-T2345$ and $NTA-N2723U$, and the mobile phone are $OPPO-Reno6$, $XiaoMi K60$ and $HuaWei-P30$.

\subsection{Performance Indicators}
\subsubsection{IoU}
\begin{figure}[!h]
	\centering
	\includegraphics[width=3 in]{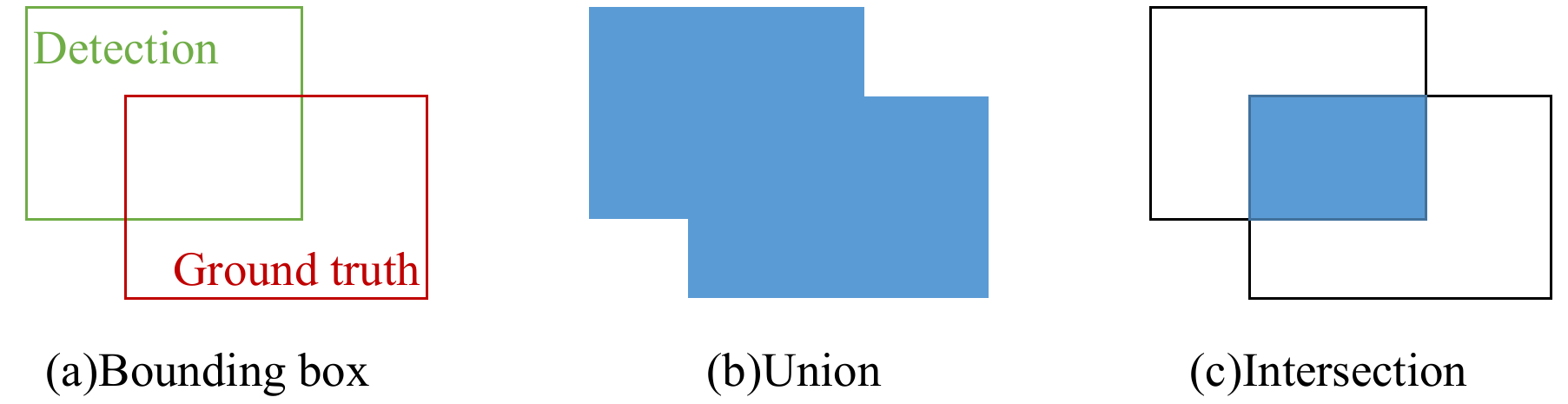}
	\caption{IoU schematic diagram.}
	\label{fig:IoU}
\end{figure}

We use intersection over union (IoU) to measure the detection accuracy, this indicator has been widely used in the object detection field. A schematic diagram is shown in Fig.~\ref{fig:IoU}. The IoU can be expressed as:

\begin{equation}
IoU=\frac{Intersection}{Union}
\label{eq:IoU}
\end{equation}

\subsubsection{Area Proportion}
\begin{figure}[!h]
 	\centering
 	\includegraphics[width=3 in]{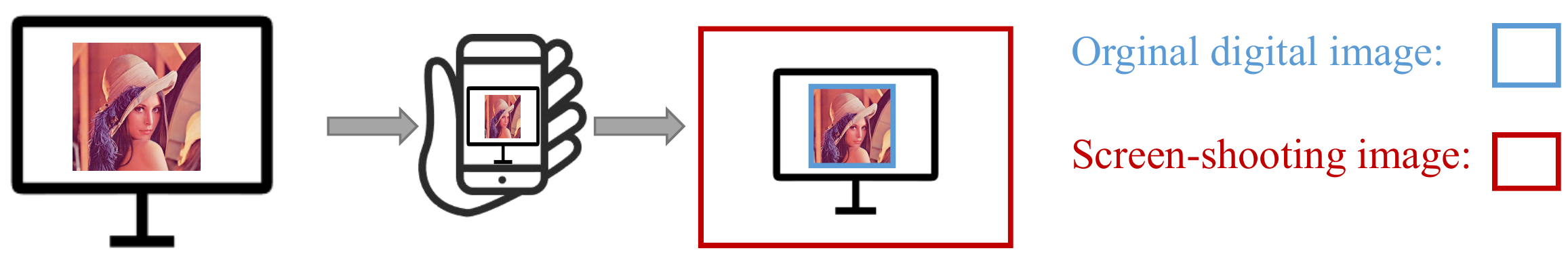}
 	\caption{Area proportion schematic diagram.}
 	\label{fig:AP}
 \end{figure}

Most methods use shooting distance as an indicator, but various hardware will make the results different greatly with a spacial shooting distance, e.g. monitor size or default camera focal length. Since we use area proportion in SSI instead distance, we keep a 4:3 shoot proportion for every experimental SSI. A schematic diagram is shown in Fig.~\ref{fig:AP}. The Area proportion can be expressed as:
\begin{equation}
Area\ Proportion=\frac{Area(ODI)}{Area(SSI)}
\label{eq:APP}
\end{equation}
where $ODI$ and $Area$ denote the area of original digital image and area finding operation respectively.

\subsubsection{Angle Offset}
The angle offset is the clamping angle between the normal line at the center of the screen and the rays line of the shooting angle.   

\subsection{The selection of embedding intensity}
\begin{figure*}[!htbp]
	\centering
	\subfigure[(a)]{\includegraphics[height=2 in]{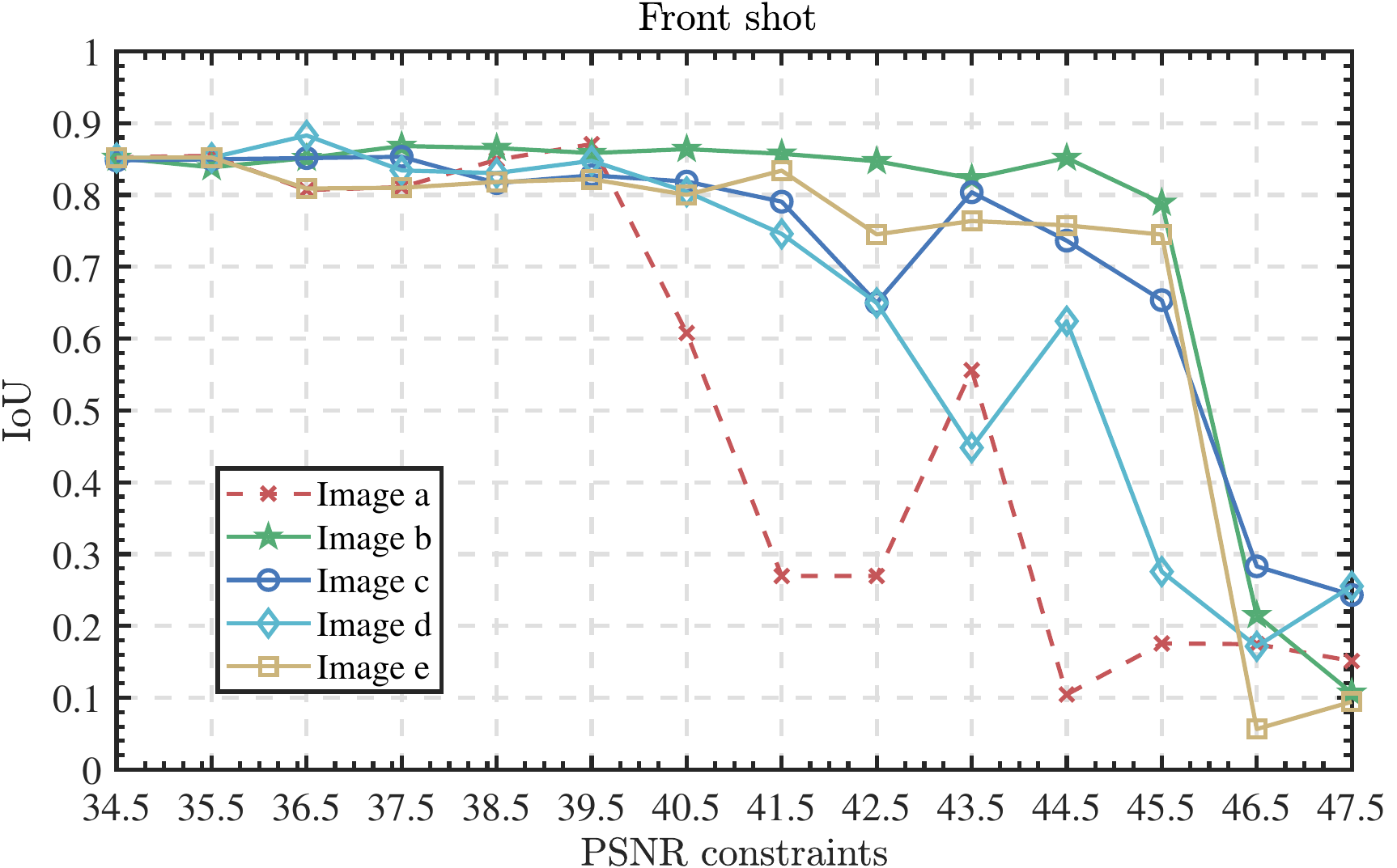}}
	\hspace{0 pt}
	\subfigure[(b)]{\includegraphics[height=2 in]{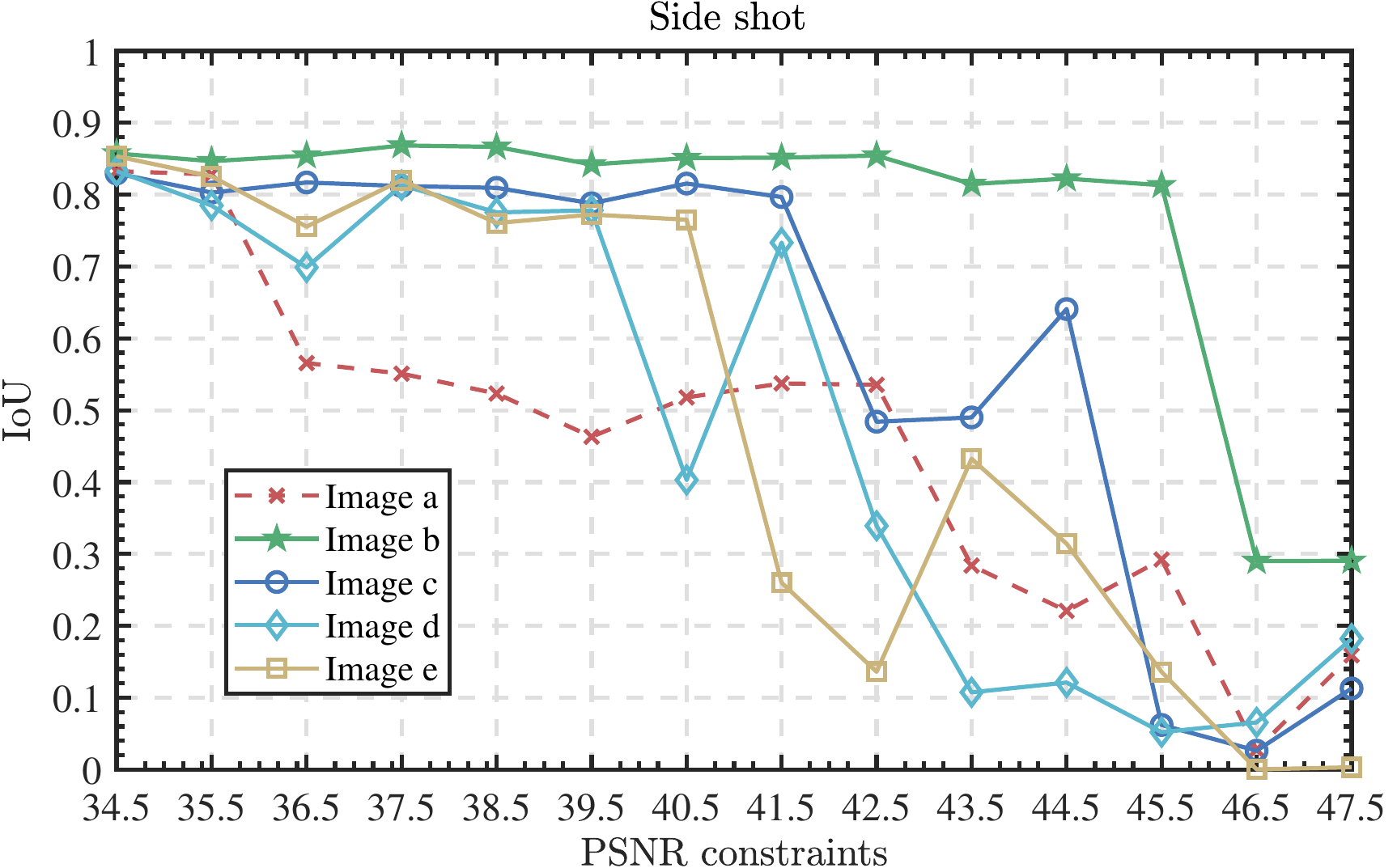}}
	\hspace{0 pt}
	\caption{The embedding intensity experiment for different image. (a) Front shot keep Area proportion between $20\%-30\%$. (b)Side shot keep angle Offset between $10\degree-20\degree$.} 
	\label{fig:intexp}
\end{figure*}

\begin{figure*}[!htbp]
	\centering
	\subfigure[]{\includegraphics[height=1.4 in]{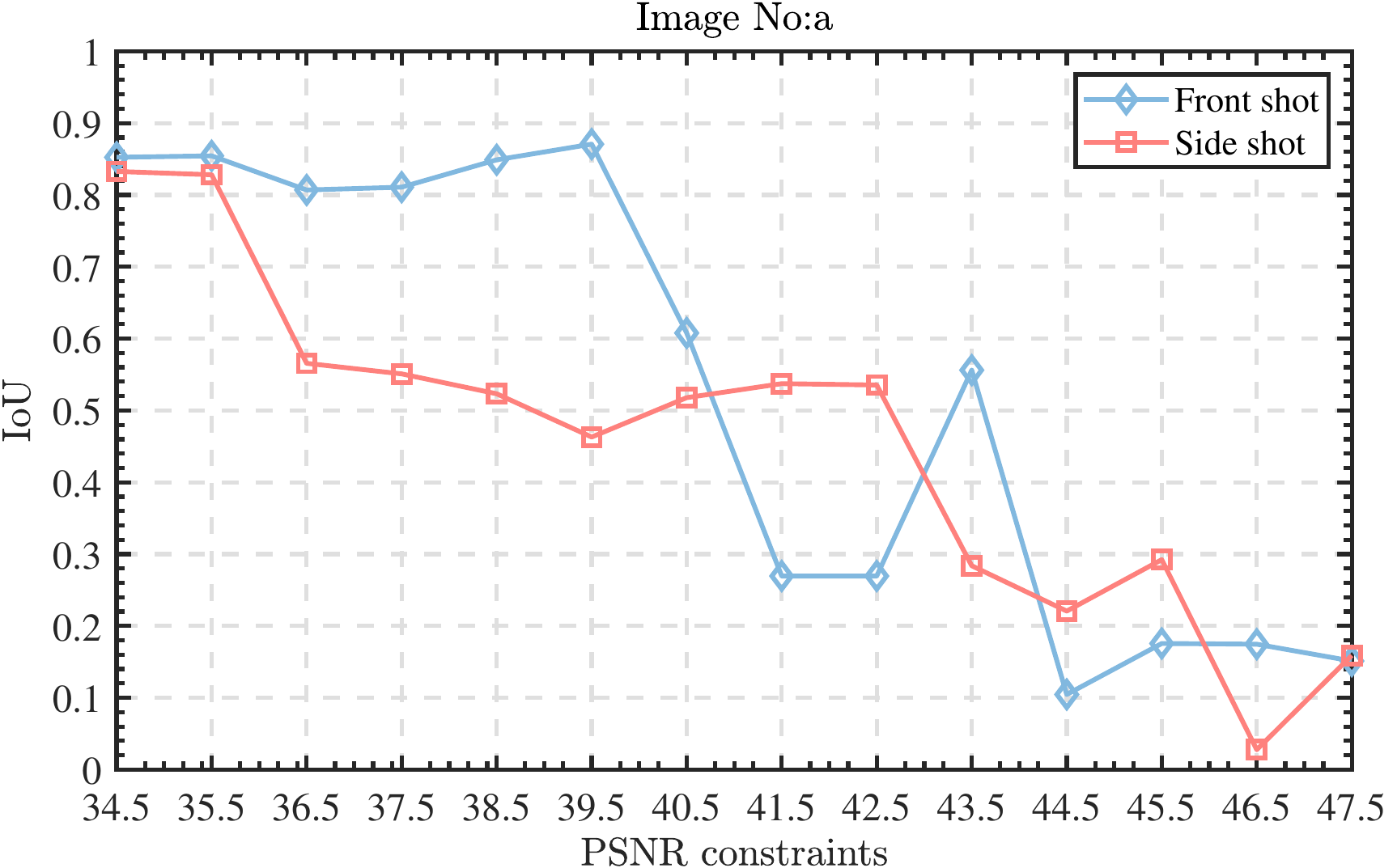}}
	\subfigure[]{\includegraphics[height=1.4 in]{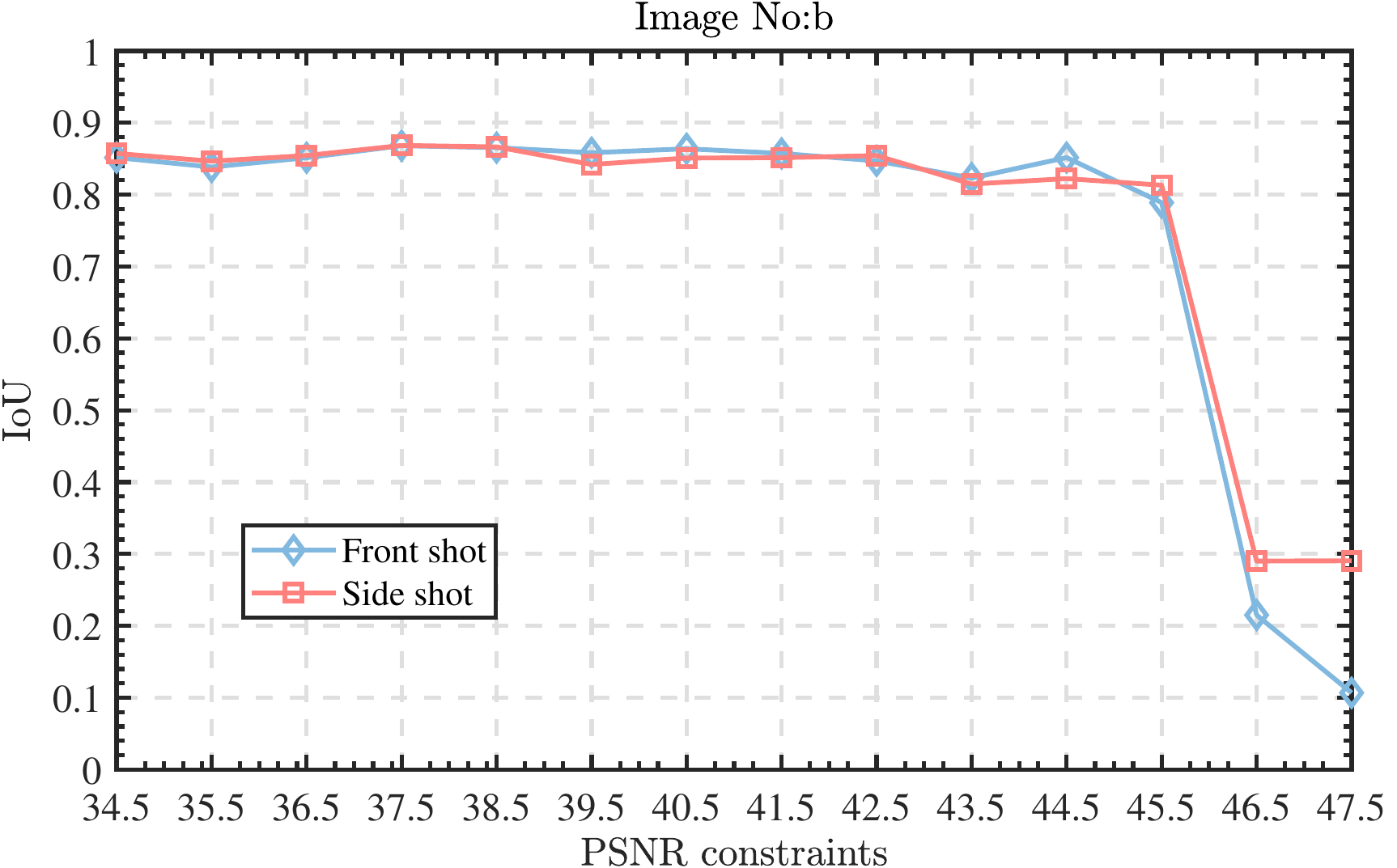}}
	\subfigure[]{\includegraphics[height=1.4 in]{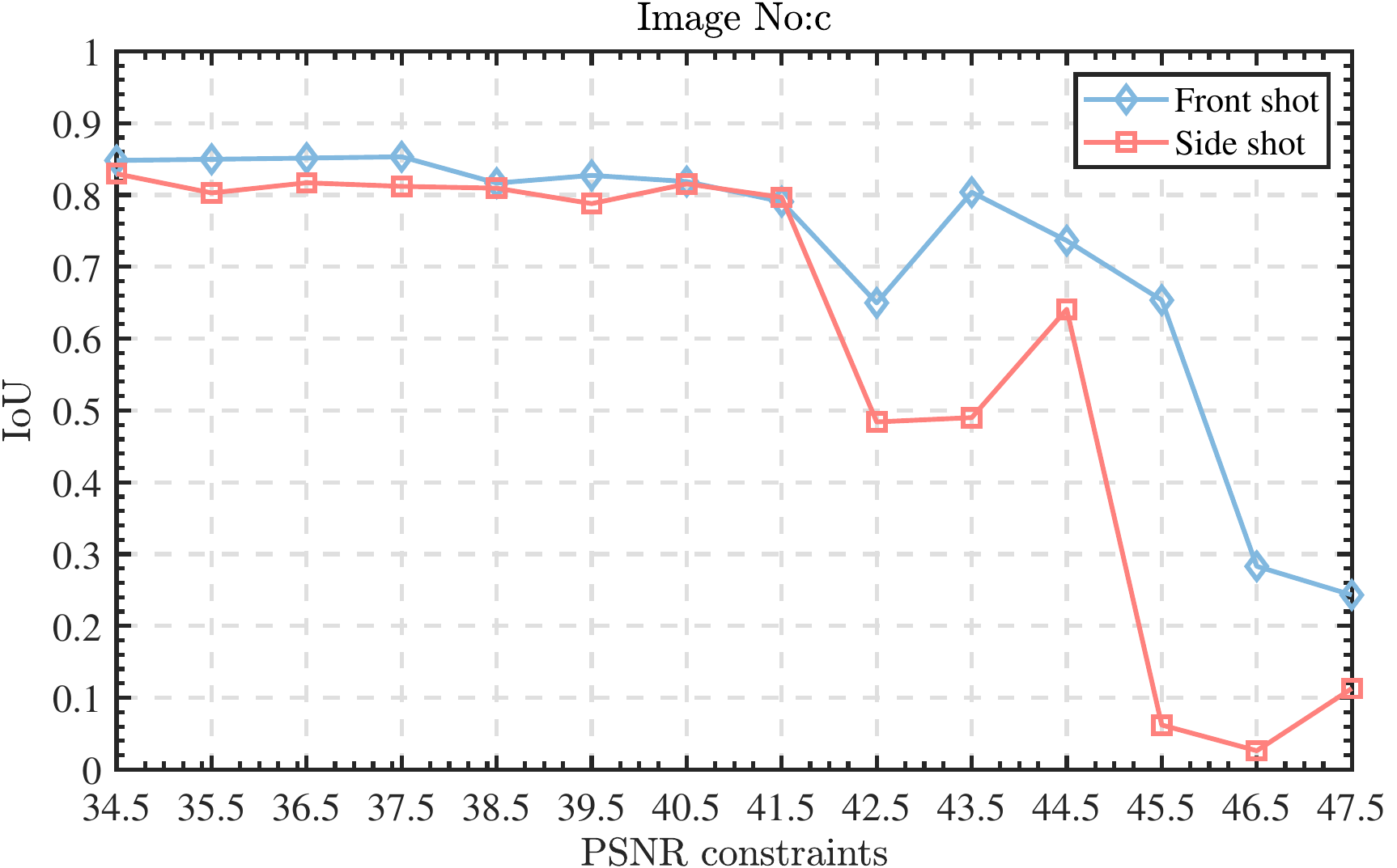}}
	\hspace{-10 pt}
	\vspace{-15 pt}
	\\
	\subfigure[]{\includegraphics[height=1.4 in]{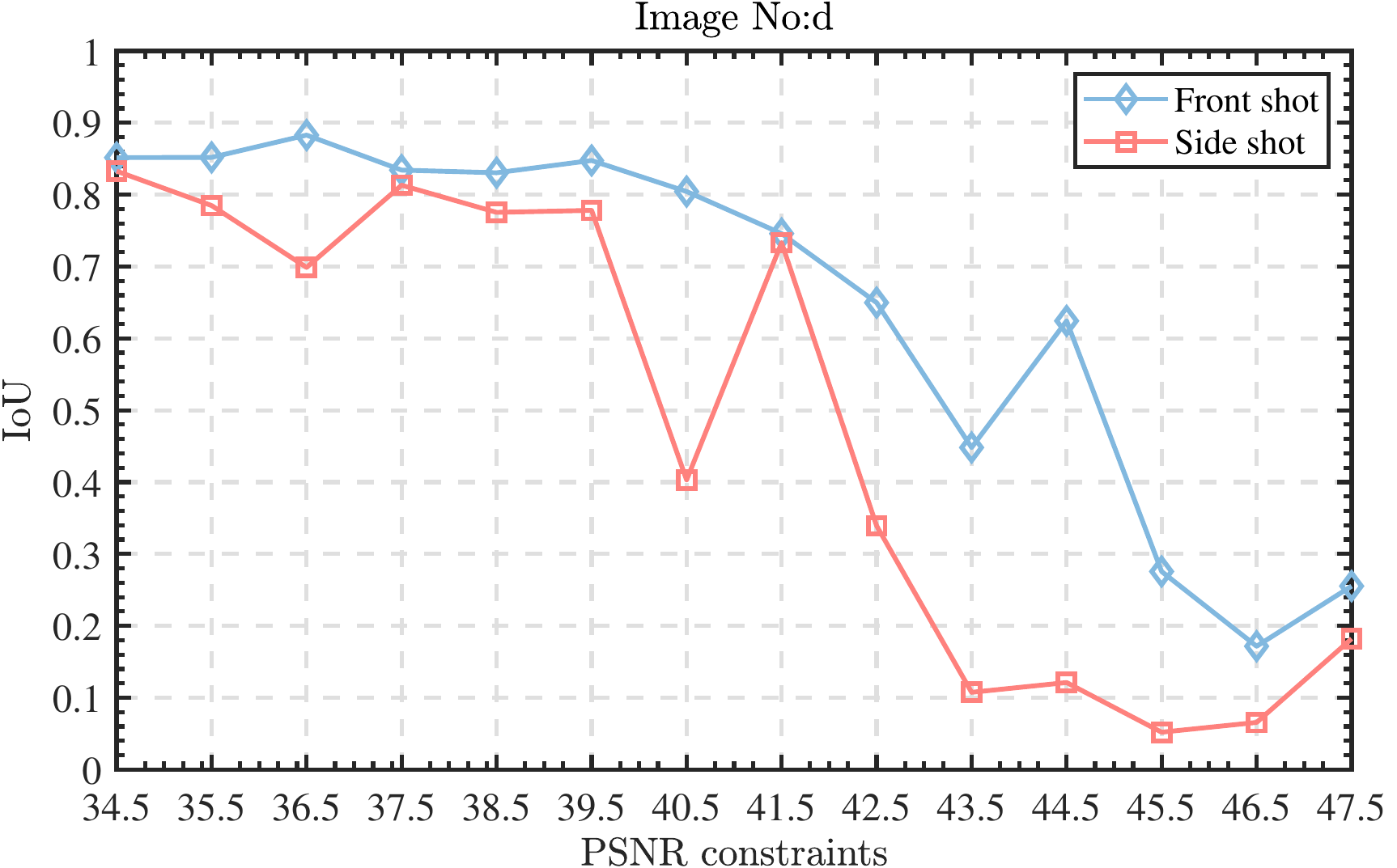}}
	\hspace{-10 pt}
	\vspace{-15 pt}
	\subfigure[]{\includegraphics[height=1.4 in]{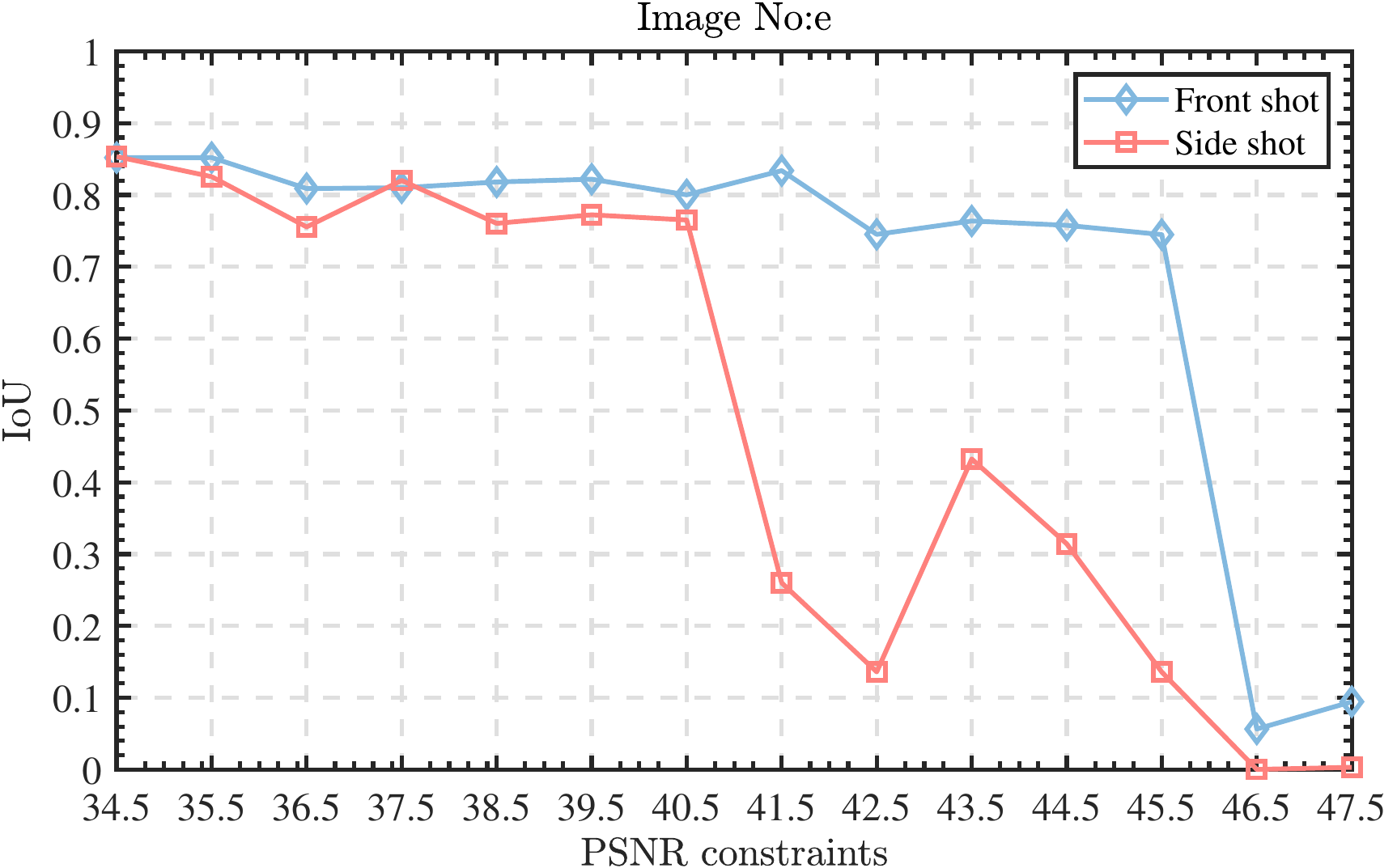}}
	\caption{The embedding intensity experiment for different shot pattern.}
	\label{fig:bexp}
\end{figure*}

The embedding intensity related to the PSNR value of marked images is discussed in \crefrange{eq:K1K2}{eq:updataK}, and it directly affects the marked image transparency. We measure the changes in the relationship between PSNR and IoU. The front shot and side shot are conducted experiments respectively. For front shot, we keep the area proportion between $20\%-30\%$, for side shot, we keep the angle offset between $10\degree-20\degree$. The experimental results are shown in Fig.~\ref{fig:intexp} and Fig.~\ref{fig:bexp}. 

In Fig.~\ref{fig:intexp}, we can see the experimental results. If $IoU > 0.8$, we can consider it an excellent detection result,
 For most of the test images, when the PSNR keeps $34.5-40.5$ the detection results are acceptable. Besides, the lower PSNR, the better IoU. Results show that the PSNR keeps 34.5, and the detection results are always excellent for a different shoot pattern.

In Fig.~\ref{fig:bexp}, we can see the experimental results in a same image. Experimental results show that the blue line is usually higher than the red line in the scheme, which means the scheme is more sensitive to angle. A possible reason is that side shots with angle affect the watermarking signals' intensity distribution.

\subsection{Area Proportion and Angle Offset experiments}
\begin{figure*}[!htbp]
	\centering
	\subfigure[]{\includegraphics[height=1.7 in]{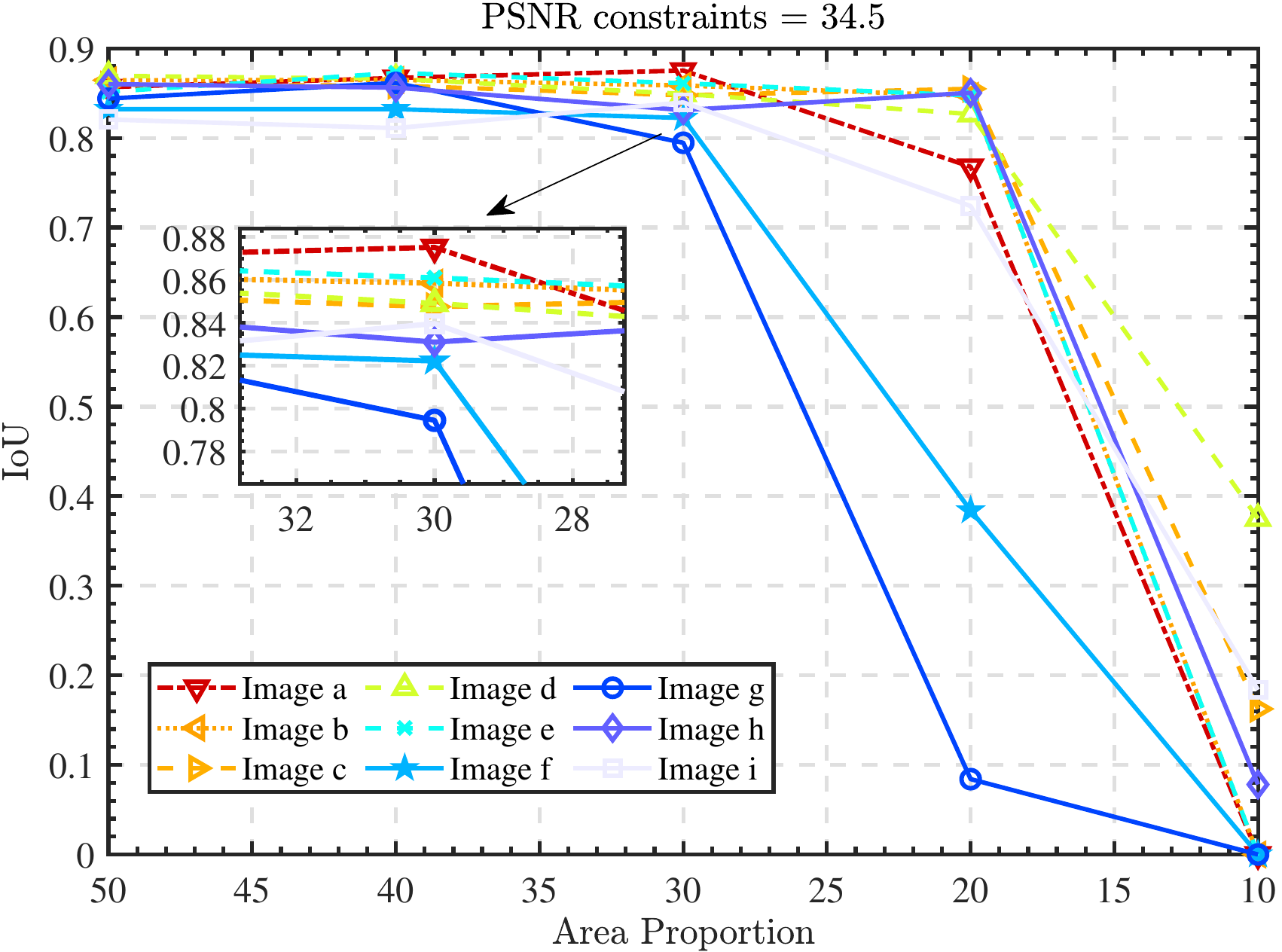}}
	\hspace{0 pt}
	\subfigure[]{\includegraphics[height=1.7 in]{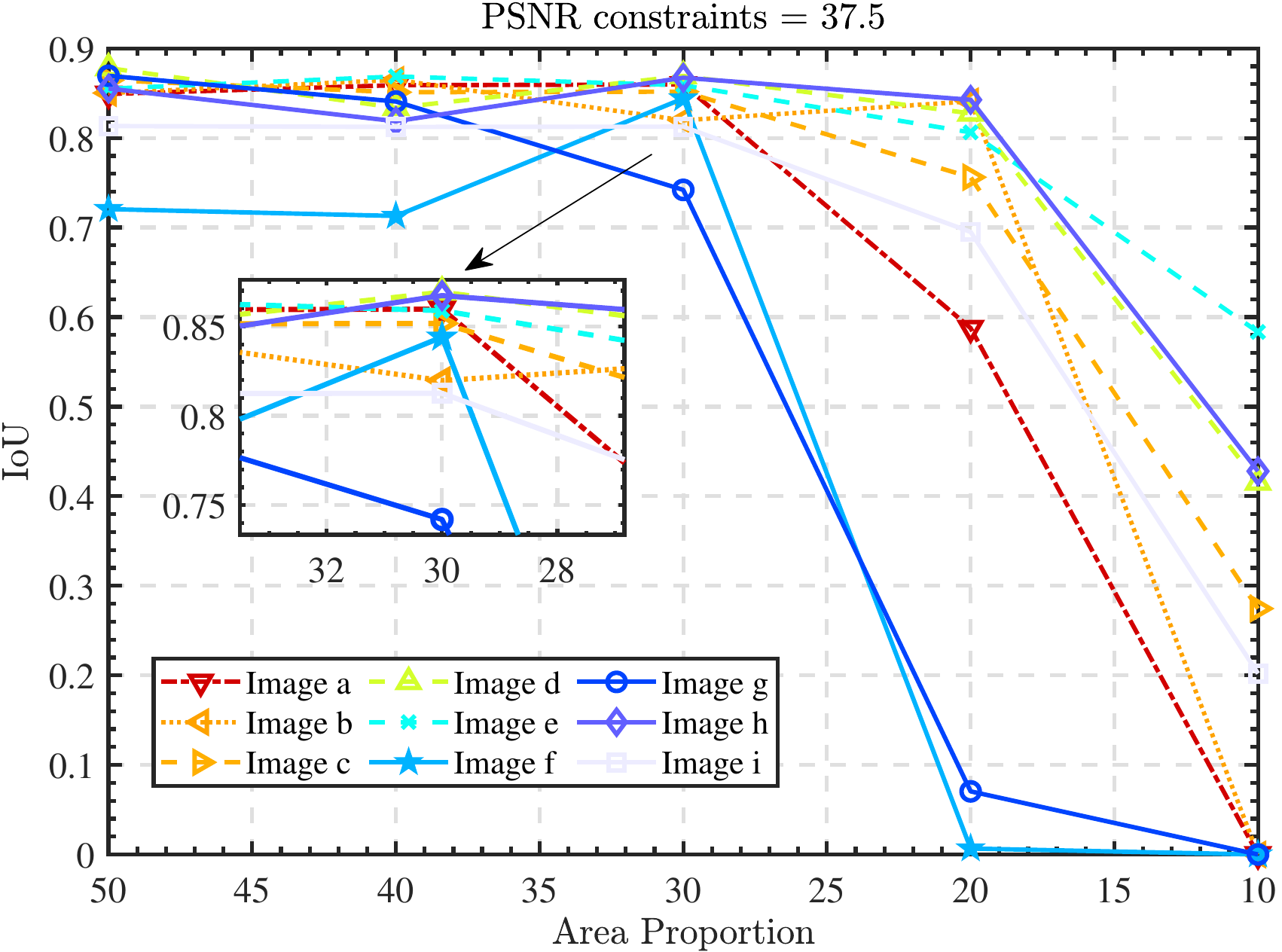}}
	\hspace{0 pt}
	\subfigure[]{\includegraphics[height=1.7 in]{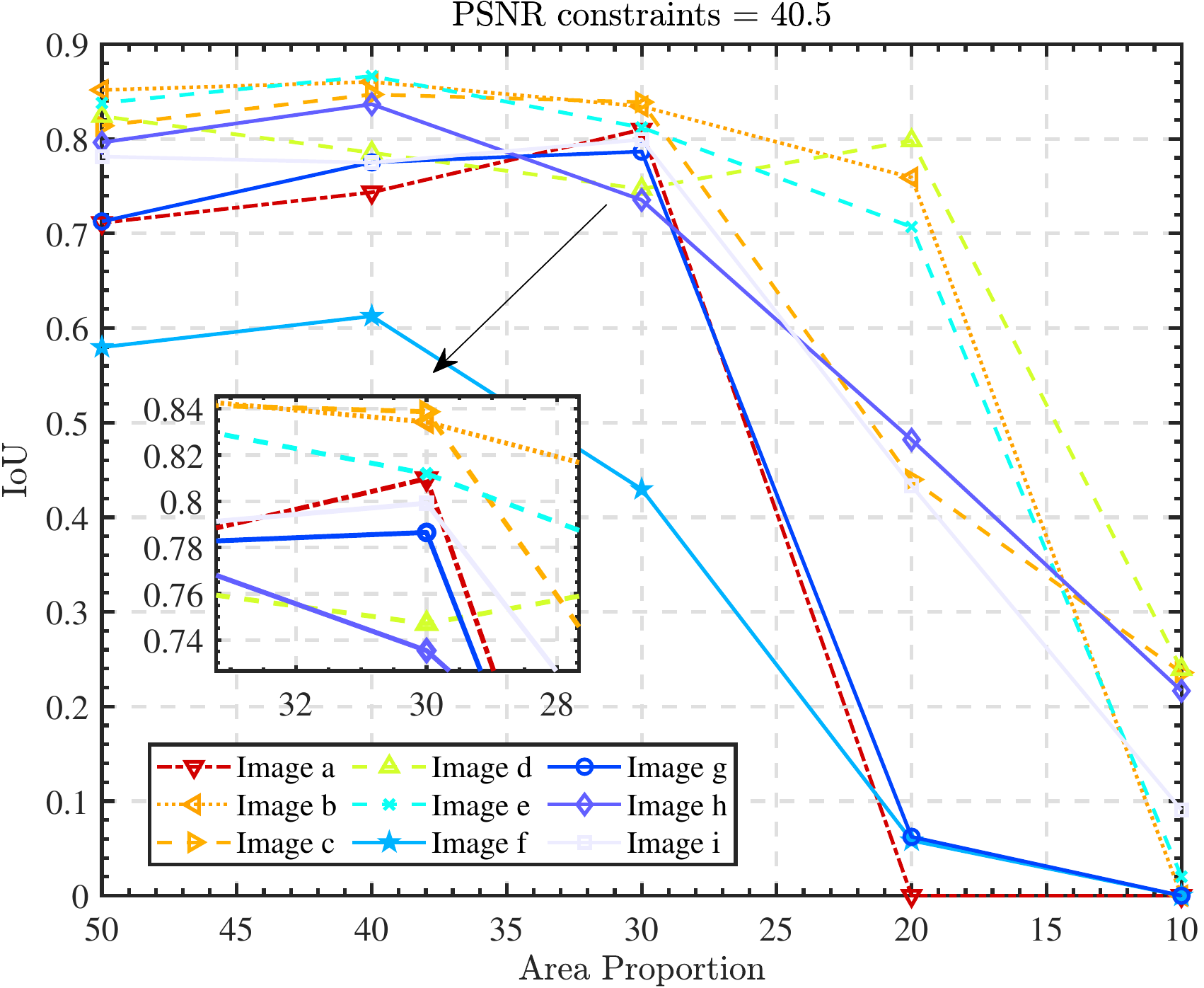}}
	\caption{Area proportion experimental results.}
	\label{fig:DEE}
\end{figure*}

\begin{figure*}[!htbp]
  	\centering
  	\subfigure[]{\includegraphics[height=1.7 in]{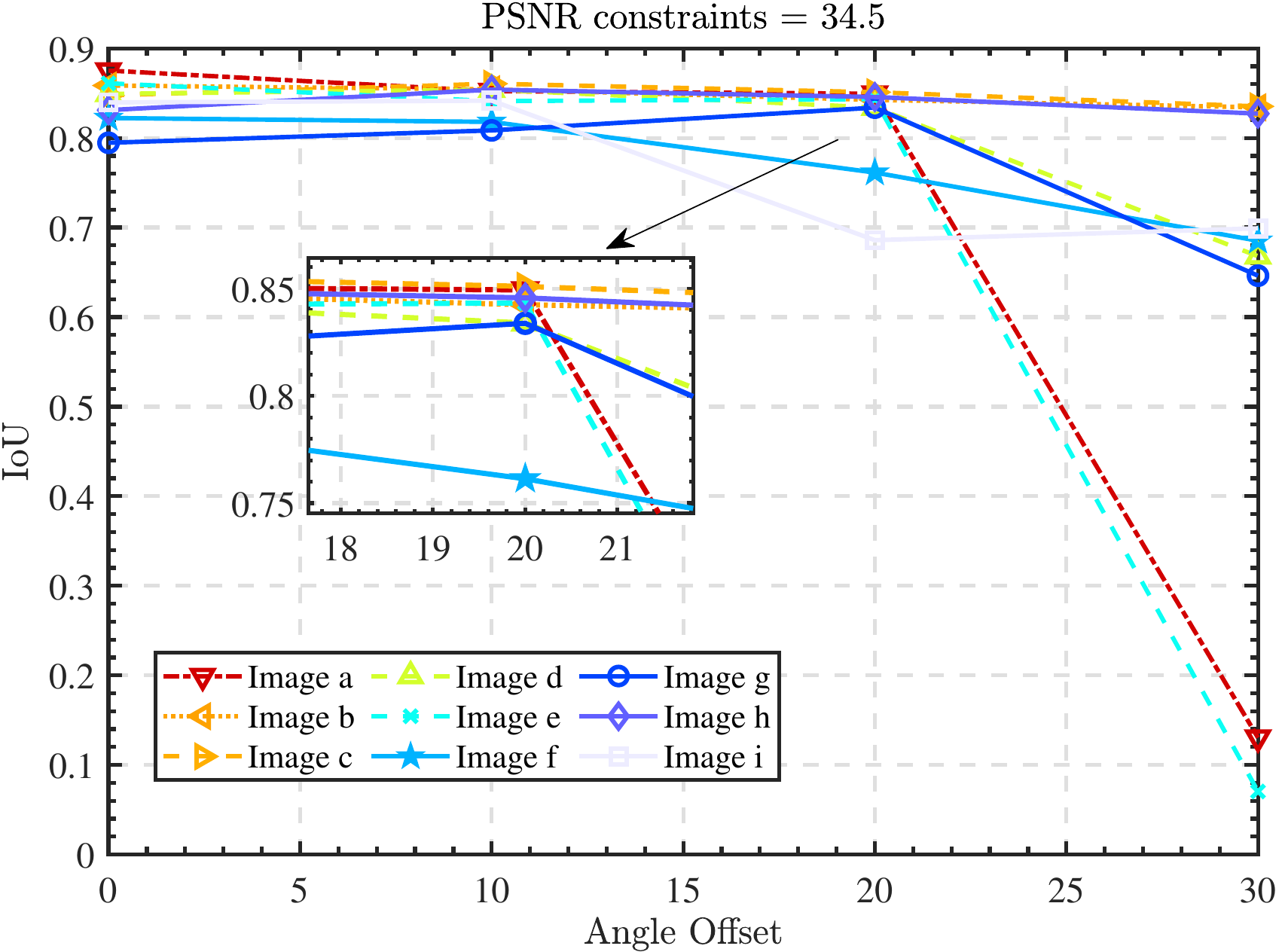}}
  	\hspace{0 pt}
  	\subfigure[]{\includegraphics[height=1.7 in]{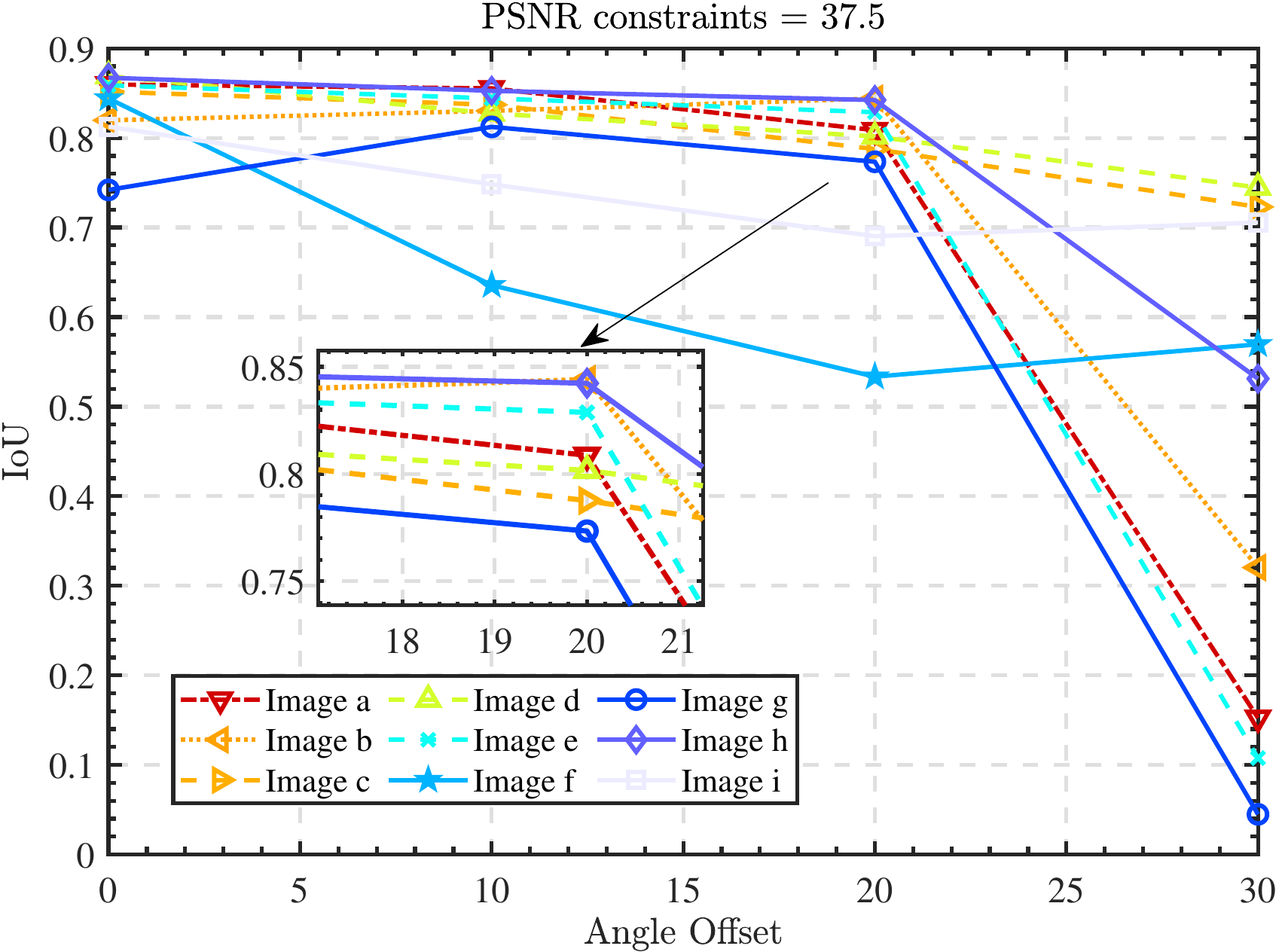}}
   	\hspace{0 pt}
  	\subfigure[]{\includegraphics[height=1.7 in]{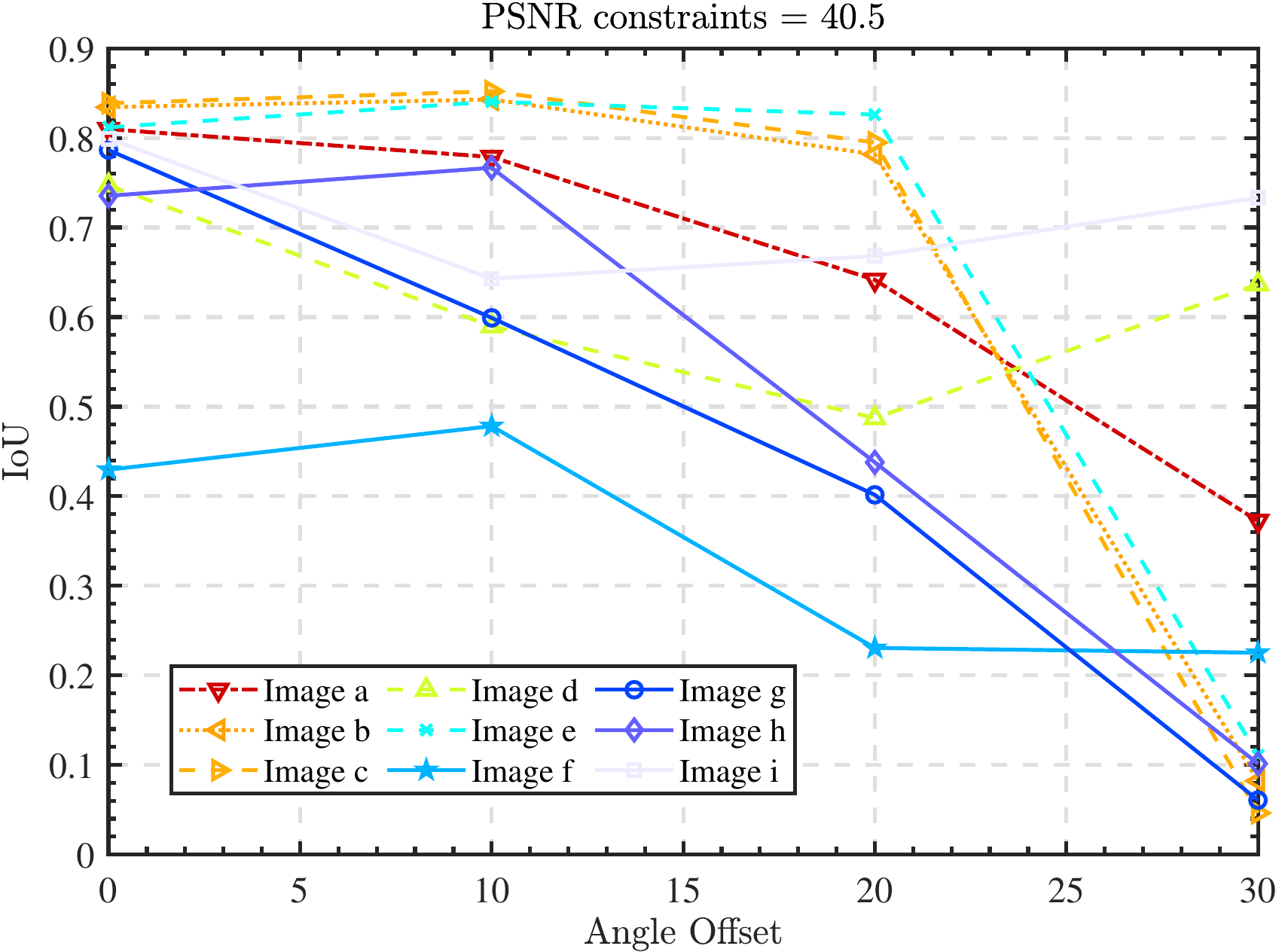}}
  	\caption{Angle Offset experimental results.}
  	\label{fig:AEE}
\end{figure*}

Further, we select embedded intensities are 34.5, 37.5, and 40.5 to conduct area proportion experiments and angle offset experiments. The experimental results are shown in Fig.~\ref{fig:DEE} and Fig.~\ref{fig:AEE}.

In Fig.~\ref{fig:DEE}, we can see the experimental results with different area proportions. For most images, when the area proportion is lower than $10\%$, the detection accuracy significantly decreases. Besides, when the area proportion is greater than $30\%$, the most of test images IoU can reach $80\%$.

In Fig.~\ref{fig:AEE}, we can see the experimental results with different angle offsets. For most image, when the Angle Offset is greater than $30\degree$, the detection accuracy significantly decrease. Besides, when angle offset is lower than $20\degree$, most of the images in PSNR constraint 34.5 and 37.5 IoU can reach $80\%$.

In addition, for area proportion experiments, we conduct another experiment to test the influence of record devices.  Table~\ref{tab:64PPP} shows high pixel record devices can get a better performance in extreme conditions. This also shows that the photo equipment is crucial to the results.
\begin{table*}[!htbp]
\centering
\caption{The IoU with different record devices in extreme conditions:Area proportion $<20\%$ }
\label{tab:64PPP}
\begin{tabular}{cclllllllllll}

\hline
\multicolumn{1}{c|}{PSNR constraint}                     & \multicolumn{4}{c|}{34.5 db}                                                                                                                                                                     & \multicolumn{4}{c|}{37.5 db}                                                                                                                                                                    & \multicolumn{4}{c}{40.5 db}                                                                                                                   \\ \hline
\multicolumn{1}{c|}{Shooting pixels} & \multicolumn{2}{c}{6400 M pixels}                                                                     & \multicolumn{2}{c}{1200 M pixels}                                                       & \multicolumn{2}{c}{6400 M pixels}                                                                     & \multicolumn{2}{c}{1200 M pixels}                                                       & \multicolumn{2}{c}{6400 M pixels}                                                       & \multicolumn{2}{c}{1200 M pixels}                   \\ \hline
\multicolumn{1}{c|}{Area proportion} & 10\%                                                        & \multicolumn{1}{c}{20\%}                & \multicolumn{1}{c}{10\%} & \multicolumn{1}{c|}{20\%}                                    & \multicolumn{1}{c}{10\%}               & \multicolumn{1}{c|}{20\%}                                    & \multicolumn{1}{c}{10\%} & \multicolumn{1}{c|}{20\%}                                    & \multicolumn{1}{c}{10\%} & \multicolumn{1}{c|}{20\%}                                    & \multicolumn{1}{c}{10\%} & \multicolumn{1}{c}{20\%} \\ \hline
\multicolumn{1}{c|}{Image:c}         & {\color[HTML]{009901} \textbf{0.83781}}                     & {\color[HTML]{009901} \textbf{0.84595}} & 0.16254                  & \multicolumn{1}{l|}{{\color[HTML]{3166FF} \textbf{0.85488}}} & 0.37952                                & \multicolumn{1}{l|}{{\color[HTML]{009901} \textbf{0.86117}}} & 0.27456                  & \multicolumn{1}{l|}{0.75615}                                 & 0.68013                  & \multicolumn{1}{l|}{{\color[HTML]{009901} \textbf{0.81846}}} & 0.23585                  & 0.43993                  \\ \hline
\multicolumn{1}{c|}{Image:e}         & \multicolumn{1}{l}{{\color[HTML]{009901} \textbf{0.81554}}} & {\color[HTML]{009901} \textbf{0.84835}} & 0                        & \multicolumn{1}{l|}{{\color[HTML]{3166FF} \textbf{0.8474}}}  & {\color[HTML]{009901} \textbf{0.8496}} & \multicolumn{1}{l|}{{\color[HTML]{009901} \textbf{0.87406}}} & 0.5834                   & \multicolumn{1}{l|}{{\color[HTML]{3531FF} \textbf{0.80624}}} & 0.41018                  & \multicolumn{1}{l|}{{\color[HTML]{009901} \textbf{0.8558}}}  & 0.020113                 & 0.70695                  \\ \hline
\multicolumn{1}{c|}{Image:h}         & \multicolumn{1}{l}{{\color[HTML]{009901} \textbf{0.85471}}} & {\color[HTML]{009901} \textbf{0.85323}} & 0.078185                 & \multicolumn{1}{l|}{{\color[HTML]{3166FF} \textbf{0.8504}}}  & 0.076469                               & \multicolumn{1}{l|}{{\color[HTML]{009901} \textbf{0.87763}}} & 0.42795                  & \multicolumn{1}{l|}{{\color[HTML]{3531FF} \textbf{0.84258}}} & 0.37279                  & \multicolumn{1}{l|}{0.68353}                                 & 0.2165                   & 0.48195                  \\ \hline
\end{tabular}
\end{table*}

\subsection{Comparison experiments}
We choose Chen's scheme \cite{chen2020screenbline} as comparison experiments. Because his scheme can not detect 1/4 region in the marked image, so we enlarged his IoU performance by $\times 4/3$ to achieve a fair comparison. The experimental results are shown in Table.~\ref{tab:AP1}~\ref{tab:AP2}~\ref{tab:AP3}~\ref{tab:AO1}~\ref{tab:AO2}~\ref{tab:AO3}.
Those experimental results show the better IoU performance we get in any embedding intensity and any shoot pattern.
\begin{table*}[]
\centering
\caption{Area proportion comparison experiments in term of IoU with PSNR constraint equal 34.5}
\label{tab:AP1}
\begin{tabular}{cc|cccccccc|}
\hline
\multicolumn{2}{c|}{PSNR constraint}                      & \multicolumn{8}{c|}{34.5}                                                                                  \\ \hline
\multicolumn{2}{c|}{Area proportion}                      & \multicolumn{2}{c}{50\%} & \multicolumn{2}{c}{40\%} & \multicolumn{2}{c}{30\%} & \multicolumn{2}{c|}{20\%} \\ \hline
\multicolumn{2}{c|}{Methods}                              & \cite{chen2020screenbline}      & Proposed     & \cite{chen2020screenbline}      & Proposed     & \cite{chen2020screenbline}      & Proposed     & \cite{chen2020screenbline}      & Proposed      \\ \hline
\multicolumn{1}{c|}{\multirow{6}{*}{\rotatebox{90}{Phone+Monitor}}} & OPPO-Reno6+ThinkVision-T2345 & 0.48479        & \textbf{0.86437}           & 0.47992        & \textbf{0.86483}           & 0.48966        & \textbf{0.85855}           & 0.12394        & \textbf{0.84698}            \\
\multicolumn{1}{c|}{}                               & OPPO-Reno6+NTA-N2723U & 0.46999        & \textbf{0.84397}
           & 0.49047        & \textbf{0.8614}           & 0.42651        & \textbf{0.7946}           & 0.3775        & \textbf{0.08416}            \\
\multicolumn{1}{c|}{}                               & HuaWei-P30+ThinkVision-T2345 & 0.34043        & \textbf{0.85947}           & 0.30812        & \textbf{0.85837}           & 0.45125        & \textbf{0.84756}           & 0.3988        & \textbf{0.85488}            \\
\multicolumn{1}{c|}{}                               & HuaWei-P30+NTA-N2723U & 0.28959        & \textbf{0.86009}           & 0.50558        & \textbf{0.85618}           & 0.38747        & \textbf{0.83108}           & 0.2118        & \textbf{0.8504}            \\
\multicolumn{1}{c|}{}                               & XiaoMiK60+ThinkVision-T2345 & 0.47437        & \textbf{0.86937}           & 0.36362        & \textbf{0.86599}           & 0.27222        & \textbf{0.84917}           & 0.48814        & \textbf{0.82665}            \\
\multicolumn{1}{c|}{}                               & XiaoMiK60+NTA-N2723U & 0.30844        & \textbf{0.85147}           & 0.48247        & \textbf{0.87264}           & 0.48857        & \textbf{0.86102}           & 0.41491        & \textbf{0.8474}            \\ \hline
\end{tabular}
\end{table*} 

\begin{table*}[]
\centering
\caption{Area proportion comparison experiments in term of IoU with PSNR constraint equal 37.5}
\label{tab:AP2}
\begin{tabular}{cc|cccccccc}
\hline
\multicolumn{2}{c|}{PSNR constraint}                                               & \multicolumn{8}{c}{37.5}                                                                                            \\ \hline
\multicolumn{2}{c|}{Area proportion}                                               & \multicolumn{2}{c}{50\%}   & \multicolumn{2}{c}{40\%}   & \multicolumn{2}{c}{30\%}    & \multicolumn{2}{c}{20\%}    \\ \hline
\multicolumn{2}{c|}{Methods}                                                       & \cite{chen2020screenbline}    & Proposed         & \cite{chen2020screenbline}    & Proposed         & \cite{chen2020screenbline}     & Proposed         & \cite{chen2020screenbline}    & Proposed          \\ \hline
\multicolumn{1}{c|}{\multirow{6}{*}{\rotatebox{90}{Phone+Monitor}}} & OPPO-Reno6+ThinkVision-T2345 & 0.24703 & \textbf{0.84909} & 0.47885 & \textbf{0.85893} & 0.33559  & \textbf{0.85967} & 0.34267 & \textbf{0.58768}  \\
\multicolumn{1}{c|}{}                               & OPPO-Reno6+NTA-N2723U        & 0.26129 & \textbf{0.86943} & 0.34136 & \textbf{0.84069} & 0.28105  & \textbf{0.742}   & 0.24698 & \textbf{0.070487} \\
\multicolumn{1}{c|}{}                               & HuaWei-P30+ThinkVision-T2345 & 0.32361 & \textbf{0.8645}  & 0.30992 & \textbf{0.85104} & 0.34144  & \textbf{0.85168} & 0.31067 & \textbf{0.75615}  \\
\multicolumn{1}{c|}{}                               & HuaWei-P30+NTA-N2723U        & 0.30461 & \textbf{0.85515} & 0.39334 & \textbf{0.81863} & 0.30884  & \textbf{0.86724} & 0.17288 & \textbf{0.84258}  \\
\multicolumn{1}{c|}{}                               & XiaoMiK60+ThinkVision-T2345  & 0.18245 & \textbf{0.87802} & 0.37287 & \textbf{0.83369} & 0.051848 & \textbf{0.86916} & 0.10184 & \textbf{0.82743}  \\
\multicolumn{1}{c|}{}                               & XiaoMiK60+NTA-N2723U         & 0.22524 & \textbf{0.80624} & 0.47185 & \textbf{0.85882} & 0.37164  & \textbf{0.86864} & 0.42915 & \textbf{0.855}    \\ \hline
\end{tabular}
\end{table*} 

\begin{table*}[]
\centering
\caption{Area proportion comparison experiments in term of IoU with PSNR constraint equal 40.5}
\label{tab:AP3}
\begin{tabular}{cc|cccccccc}
\hline
\multicolumn{2}{c|}{PSNR constraint}                                               & \multicolumn{8}{c}{40.5}                                                                                             \\ \hline
\multicolumn{2}{c|}{Area proportion}                                               & \multicolumn{2}{c}{50\%}   & \multicolumn{2}{c}{40\%}   & \multicolumn{2}{c}{30\%}    & \multicolumn{2}{c}{20\%}     \\ \hline
\multicolumn{2}{c|}{Methods}                                                       & \cite{chen2020screenbline}    & Proposed         & \cite{chen2020screenbline}    & Proposed         & \cite{chen2020screenbline}     & Proposed         & \cite{chen2020screenbline}     & Proposed          \\ \hline
\multicolumn{1}{c|}{\multirow{6}{*}{\rotatebox{90}{Phone+Monitor}}} & OPPO-Reno6+ThinkVision-T2345 & 0.18828 & \textbf{0.85161} & 0.28578 & \textbf{0.86018} & 0.26539  & \textbf{0.83433} & 0.23375  & \textbf{0.75913}  \\
\multicolumn{1}{c|}{}                               & OPPO-Reno6+NTA-N2723U        & 0.19743 & \textbf{0.78653} & 0.21692 & \textbf{0.77514} & 0        & \textbf{0.71269} & 0        & \textbf{0.062245} \\
\multicolumn{1}{c|}{}                               & HuaWei-P30+ThinkVision-T2345 & 0.21487 & \textbf{0.81386} & 0.28779 & \textbf{0.84689} & 0.22529  & \textbf{0.83879} & 0.23277  & \textbf{0.43993}  \\
\multicolumn{1}{c|}{}                               & HuaWei-P30+NTA-N2723U        & 0.10091 & \textbf{0.79619} & 0.18335 & \textbf{0.83659} & 0.16806  & \textbf{0.73539} & 0.12289  & \textbf{0.48195}  \\
\multicolumn{1}{c|}{}                               & XiaoMiK60+ThinkVision-T2345  & 0.1519  & \textbf{0.87802} & 0.11283 & \textbf{0.83369} & 0.051823 & \textbf{0.86916} & 0.048995 & \textbf{0.82743}  \\
\multicolumn{1}{c|}{}                               & XiaoMiK60+NTA-N2723U         & 0.41696 & \textbf{0.83789} & 0.3654  & \textbf{0.86633} & 0.05295  & \textbf{0.81194} & 0.24016  & \textbf{0.70695}  \\ \hline
\end{tabular}
\end{table*}

\begin{table*}[]
\centering
\caption{Angle offset comparison experiments in term of IoU with PSNR constraint equal 34.5}
\label{tab:AO1}
\begin{tabular}{cc|llllllll}
\hline
\multicolumn{2}{c|}{PSNR constraint}                                               & \multicolumn{8}{c}{34.5}                                                                                                                                                                                                              \\ \hline
\multicolumn{2}{c|}{Area proportion}                                               & \multicolumn{2}{c}{0°}                                  & \multicolumn{2}{c}{10°}                                 & \multicolumn{2}{c}{20°}                                 & \multicolumn{2}{c}{30°}                                 \\ \hline
\multicolumn{2}{c|}{Methods}                                                       & \multicolumn{1}{c}{\cite{chen2020screenbline}} & \multicolumn{1}{c}{Proposed} & \multicolumn{1}{c}{\cite{chen2020screenbline}} & \multicolumn{1}{c}{Proposed} & \multicolumn{1}{c}{\cite{chen2020screenbline}} & \multicolumn{1}{c}{Proposed} & \multicolumn{1}{c}{\cite{chen2020screenbline}} & \multicolumn{1}{c}{Proposed} \\ \hline
\multicolumn{1}{c|}{\multirow{6}{*}{\rotatebox{90}{Phone+Monitor}}} & OPPO-Reno6+ThinkVision-T2345 & 0.4276                   & \textbf{0.87525}             & 0.4276                   & \textbf{0.85228}             & 0.34253                  & \textbf{0.84939}             & \textbf{0.30994}         & 0.13047                      \\
\multicolumn{1}{c|}{}                               & OPPO-Reno6+NTA-N2723U        & 0.49047                  & \textbf{0.7946}              & 0.27502                  & \textbf{0.80843}             & 0.30174                  & \textbf{0.83386}             & 0.22391                  & \textbf{0.64622}             \\
\multicolumn{1}{c|}{}                               & HuaWei-P30+ThinkVision-T2345 & 0.30812                  & \textbf{0.84756}             & 0.30144                  & \textbf{0.86057}             & 0.23879                  & \textbf{0.85105}             & 0.25453                  & \textbf{0.83524}             \\
\multicolumn{1}{c|}{}                               & HuaWei-P30+NTA-N2723U        & 0.25831                  & \textbf{0.8397}              & 0.10782                  & \textbf{0.84163}             & 0.098753                 & \textbf{0.68588}             & 0.17752                  & \textbf{0.69911}             \\
\multicolumn{1}{c|}{}                               & XiaoMiK60+ThinkVision-T2345  & 0.36362                  & \textbf{0.84917}             & 0.30448                  & \textbf{0.85395}             & 0.20628                  & \textbf{0.834}               & 0.28088                  & \textbf{0.66794}             \\
\multicolumn{1}{c|}{}                               & XiaoMiK60+NTA-N2723U         & 0.48247                  & \textbf{0.86102}             & 0.4012                   & \textbf{0.84115}             & 0.28648                  & \textbf{0.84331}             & 0.39225                  & \textbf{0.70364}             \\ \hline
\end{tabular}
\end{table*}

\begin{table*}[]
\centering
\caption{Angle offset comparison experiments in term of IoU with PSNR constraint equal 37.5}
\label{tab:AO2}
\begin{tabular}{cc|cccccccc}
\hline
\multicolumn{2}{c|}{PSNR constraint}                                               & \multicolumn{8}{c}{37.5}                                                                                                    \\ \hline
\multicolumn{2}{c|}{Area proportion}                                               & \multicolumn{2}{c}{0°}     & \multicolumn{2}{c}{10°}    & \multicolumn{2}{c}{20°}     & \multicolumn{2}{c}{30°}             \\ \hline
\multicolumn{2}{c|}{Methods}                                                       & \cite{chen2020screenbline}    & Proposed         & \cite{chen2020screenbline}    & Proposed         & \cite{chen2020screenbline}     & Proposed         & \cite{chen2020screenbline}             & Proposed         \\ \hline
\multicolumn{1}{c|}{\multirow{6}{*}{\rotatebox{90}{Phone+Monitor}}} & OPPO-Reno6+ThinkVision-T2345 & 0.33559 & \textbf{0.85967} & 0.2712  & \textbf{0.85533} & 0.2668   & \textbf{0.8087}  & \textbf{0.26695} & 0.15249          \\
\multicolumn{1}{c|}{}                               & OPPO-Reno6+NTA-N2723U        & 0.28105 & \textbf{0.742}   & 0.23426 & \textbf{0.81234} & 0.12567  & \textbf{0.7734}  & \textbf{0.28076} & 0.04484          \\
\multicolumn{1}{c|}{}                               & HuaWei-P30+ThinkVision-T2345 & 0.34144 & \textbf{0.85168} & 0.2333  & \textbf{0.8371}  & 0.15468  & \textbf{0.78759} & 0.08884          & \textbf{0.72299} \\
\multicolumn{1}{c|}{}                               & HuaWei-P30+NTA-N2723U        & 0.2897  & \textbf{0.8126}  & 0.10563 & \textbf{0.74816} & 0.09071 & \textbf{0.69027} & 0                & \textbf{0.70554} \\
\multicolumn{1}{c|}{}                               & XiaoMiK60+ThinkVision-T2345  & 0.05185 & \textbf{0.86916} & 0.29832 & \textbf{0.82734} & 0.04541  & \textbf{0.80168} & 0.1539           & \textbf{0.74464} \\
\multicolumn{1}{c|}{}                               & XiaoMiK60+NTA-N2723U         & 0.37164 & \textbf{0.85882} & 0.37487 & \textbf{0.84416} & 0.23368  & \textbf{0.82871} & \textbf{0.35813} & 0.10764          \\ \hline
\end{tabular}
\end{table*}

\begin{table*}[]
\centering
\caption{Angle offset comparison experiments in term of IoU with PSNR constraint equal 40.5}
\label{tab:AO3}
\begin{tabular}{cc|cccccccc}
\hline
\multicolumn{2}{c|}{PSNR constraint}                                               & \multicolumn{8}{c}{40.5}                                                                                                    \\ \hline
\multicolumn{2}{c|}{Area proportion}                                               & \multicolumn{2}{c}{0°}      & \multicolumn{2}{c}{10°}    & \multicolumn{2}{c}{20°}    & \multicolumn{2}{c}{30°}             \\ \hline
\multicolumn{2}{c|}{Methods}                                                       & \cite{chen2020screenbline}     & Proposed         & \cite{chen2020screenbline}    & Proposed         & \cite{chen2020screenbline}    & Proposed         & \cite{chen2020screenbline}             & Proposed         \\ \hline
\multicolumn{1}{c|}{\multirow{6}{*}{\rotatebox{90}{Phone+Monitor}}} & OPPO-Reno6+ThinkVision-T2345 & 0.05907  & \textbf{0.80982} & 0.25447 & \textbf{0.77875} & 0.1994  & \textbf{0.64158} & 0.11813          & \textbf{0.37238} \\
\multicolumn{1}{c|}{}                               & OPPO-Reno6+NTA-N2723U        & 0        & \textbf{0.78653} & 0.21492 & \textbf{0.59913} & 0       & \textbf{0.40154} & \textbf{0.13267} & 0.06071          \\
\multicolumn{1}{c|}{}                               & HuaWei-P30+ThinkVision-T2345 & 0.22529  & \textbf{0.83879} & 0.22529 & \textbf{0.85202} & 0.22529 & \textbf{0.7949}  & \textbf{0.16771} & 0.04635          \\
\multicolumn{1}{c|}{}                               & HuaWei-P30+NTA-N2723U        & 0        & \textbf{0.79919} & 0       & \textbf{0.64287} & 0.22729 & \textbf{0.66826} & 0                & \textbf{0.73335} \\
\multicolumn{1}{c|}{}                               & XiaoMiK60+ThinkVision-T2345  & 0.05182  & \textbf{0.74702} & 0.1378  & \textbf{0.58975} & 0       & \textbf{0.48723} & 0.05338          & \textbf{0.63668} \\
\multicolumn{1}{c|}{}                               & XiaoMiK60+NTA-N2723U         & 0.05295 & \textbf{0.81194} & 0.30591 & \textbf{0.84049} & 0.17426 & \textbf{0.82602} & \textbf{0.21935} & 0.1107           \\ \hline
\end{tabular}
\end{table*}

\subsection{Detection in various complex scenarios}
\begin{figure*}[!htbp]
	\centering
	\subfigure[(a)Long distance]{\includegraphics[height=1.5 in]{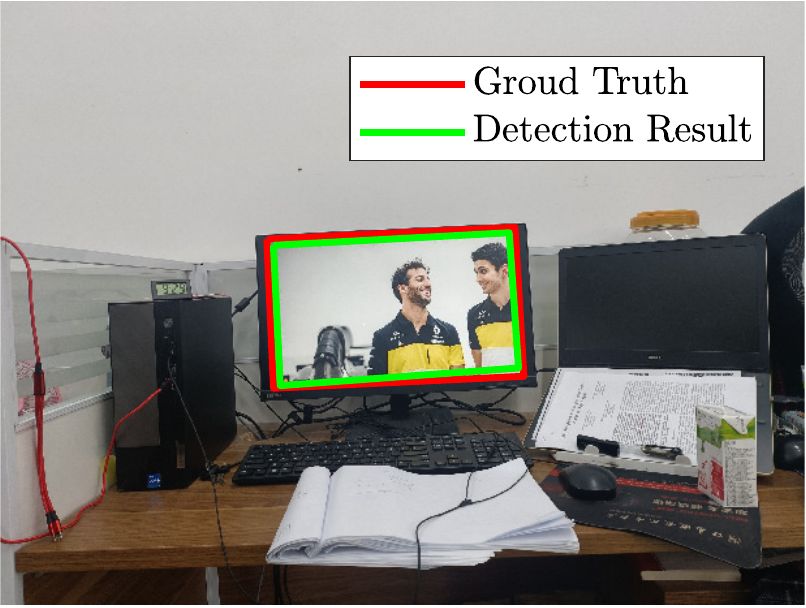}}
	\hspace{0 pt}
	\vspace{-8 pt}
	\subfigure[(b)Big angle]{\includegraphics[height=1.5 in]{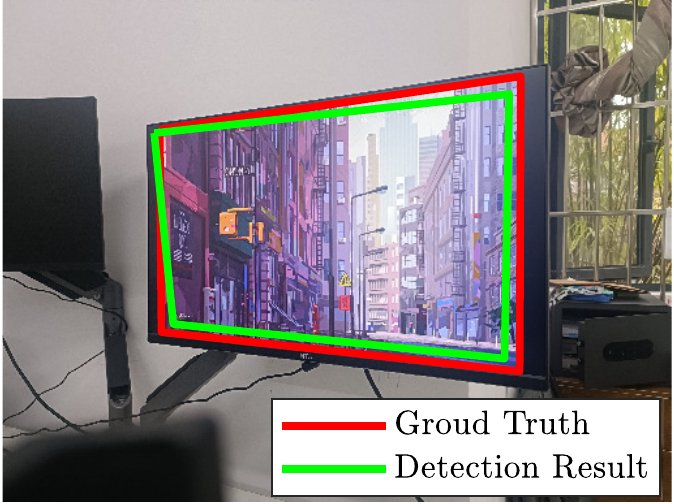}}
	\hspace{0 pt}
	\subfigure[(c)Curved Screen]{\includegraphics[height=1.5 in]{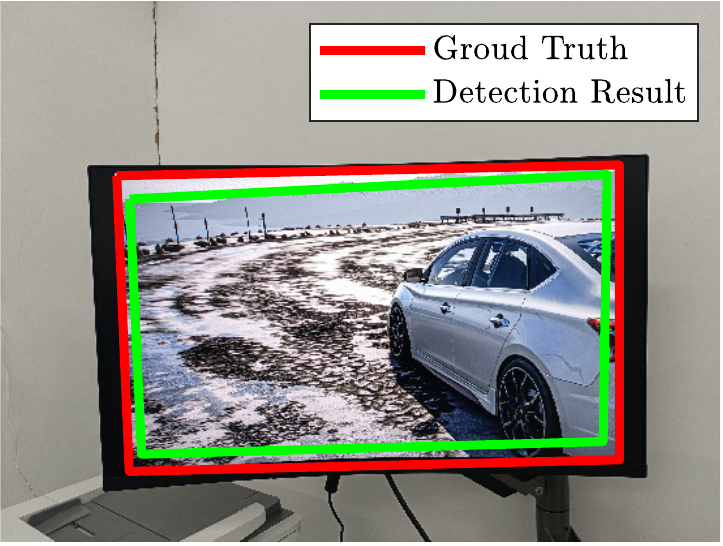}}
	\hspace{0 pt}
	\\
	\subfigure[(d)Print-camera]{\includegraphics[height=1.5 in]{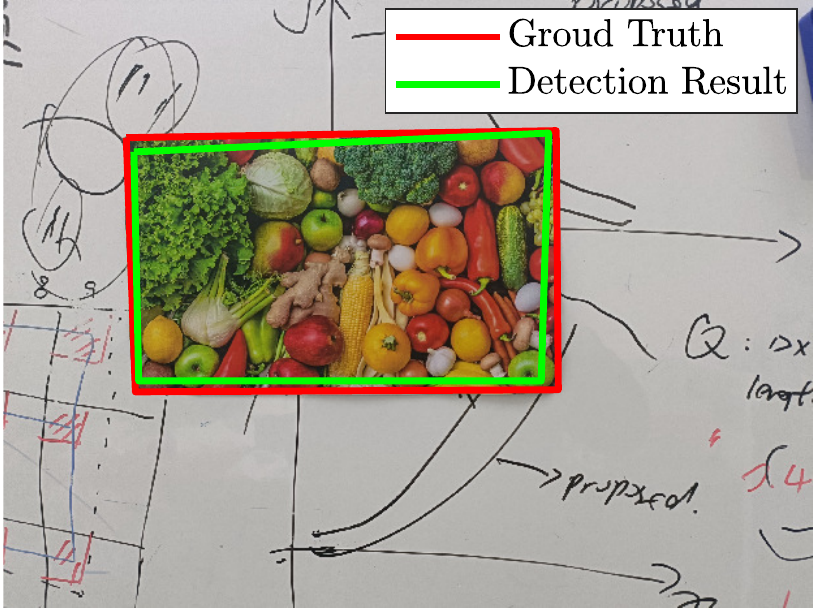}}
	\hspace{0 pt}
	\vspace{-8 pt}
	\subfigure[(e)Print-crop-camera]{\includegraphics[height=1.5 in]{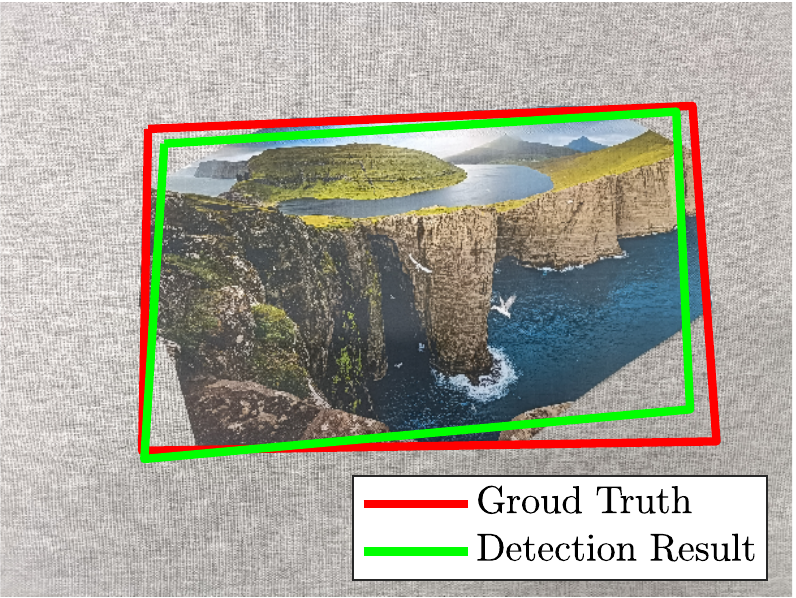}}
	\hspace{0 pt}
	\subfigure[(f)Illumination changed A]{\includegraphics[height=1.5 in]{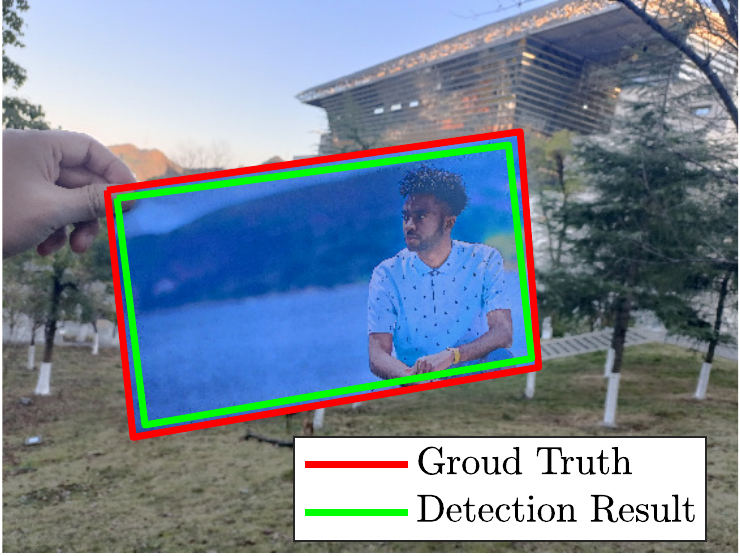}}
	\hspace{0 pt}
	\\
	\subfigure[(g)Illumination changed B]{\includegraphics[height=1.5 in]{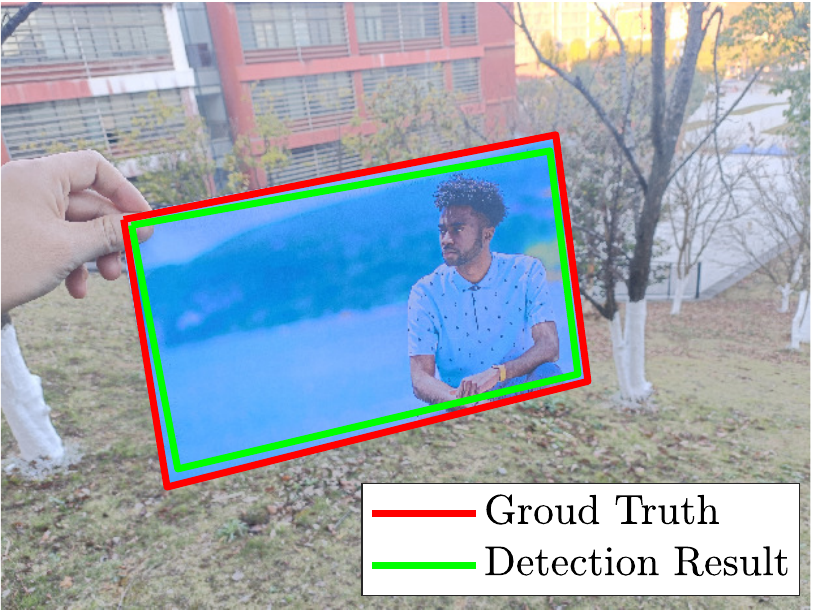}}
	\hspace{0 pt}
	\vspace{-8 pt}
	\subfigure[(h)rgb2gray]{\includegraphics[height=1.5 in]{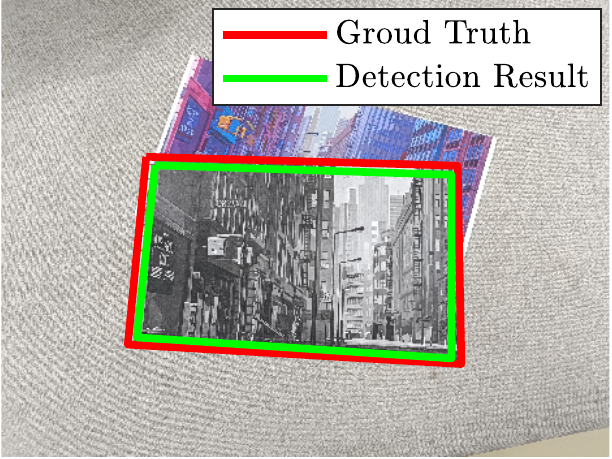}}
	\hspace{0 pt}
	\subfigure[(i)re-shooting]{\includegraphics[height=1.5 in]{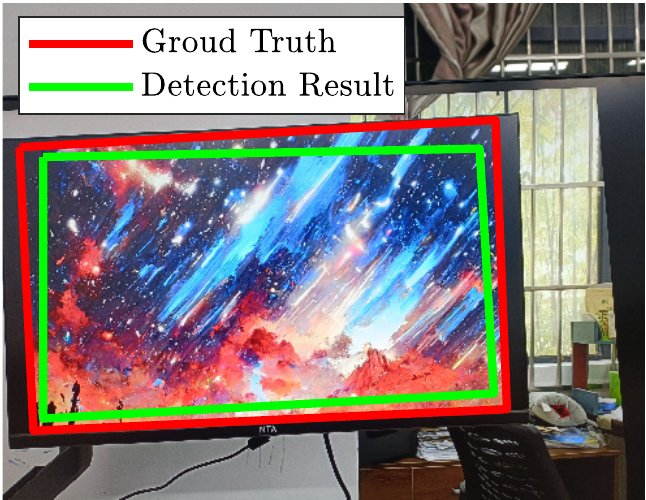}}
	\caption{Detection SSI-IRD in various complex scenarios. The IoU is (a)-0.8979,(b)-0.8673
,(c)-0.83341,(d)-0.88434,(e)-0.87577,(f)-0.86686,(g)-0.87168,(h)-0.87365,(i)-0.82492 respectively}
	\label{fig:complex}
\end{figure*}

We show some detection results in complex scenarios. The experimental results are shown in Fig.~\ref{fig:complex}. We try 9 category scenarios and shot our marked image with PSNR constraint $= 34.5$. For screen shooting scenarios, excluding area proportion Fig.~\ref{fig:complex}a and angle offset Fig.~\ref{fig:complex}b, we also try to detect in curved screen, and it's shown in Fig.~\ref{fig:complex}c. Besides, we also make experiments in Print-camera, they are shown in Fig.~\ref{fig:complex}d. For some complex scenarios in print-camera, as shown in Fig.~\ref{fig:complex}e, we crop the four corners of SSI and then detect the location. Further, two illuminations changed situations are conducted in Fig.~\ref{fig:complex}f and j, and grayscale print is detected in the experiments as Fig.~\ref{fig:complex}h. re-shooting conditions is shown in Fig.~\ref{fig:complex}i, where we shoot a SSI again, and detected re-shooting SSI in same technique. Those experimental results show excellent detection accuracy in complex scenarios.

\subsection{Robustness for AWD}
\begin{figure*}[!h]
	\centering
	\subfigure[]{\includegraphics[height=2 in]{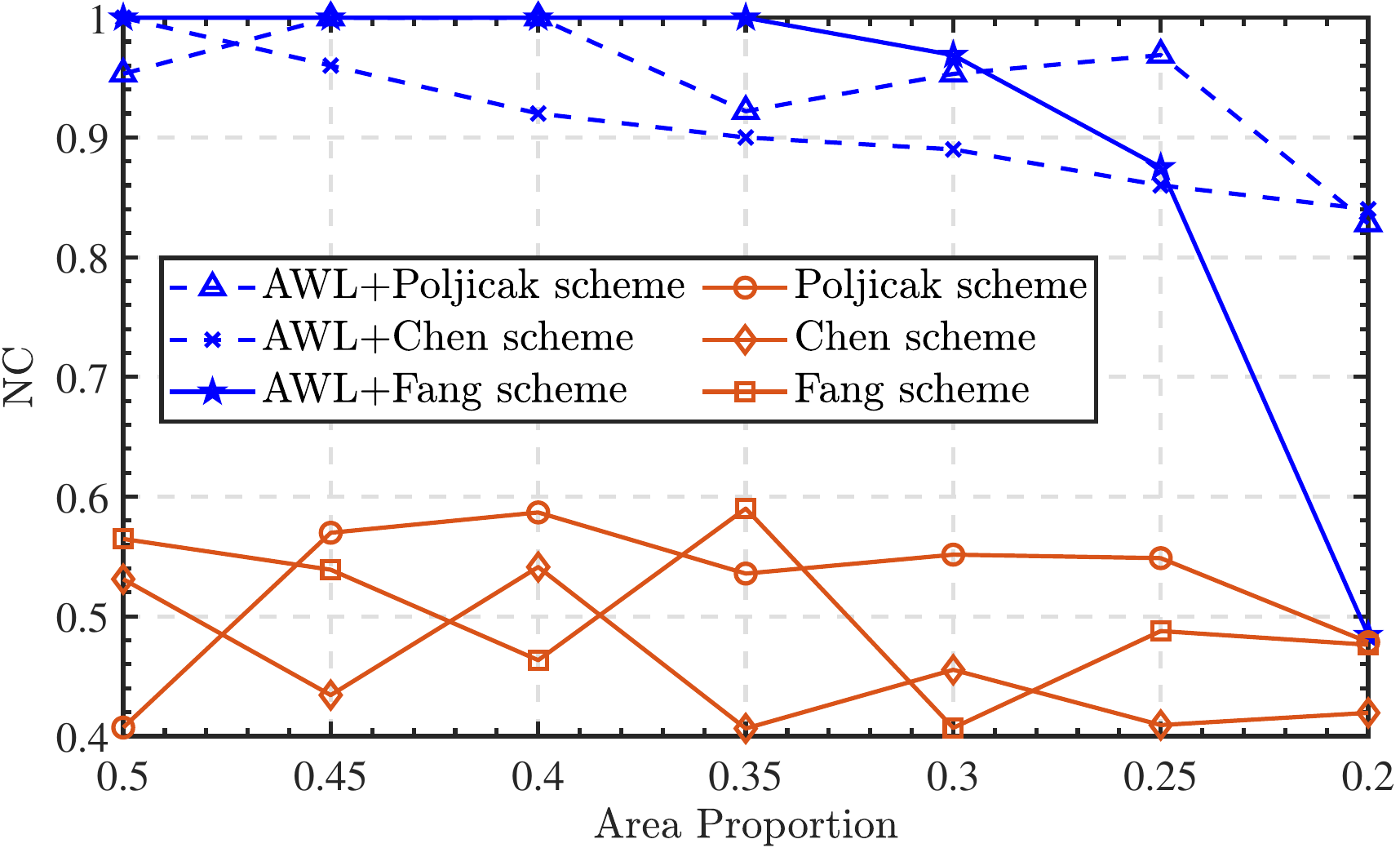}}
	\hspace{0 pt}
	\subfigure[]{\includegraphics[height=2 in]{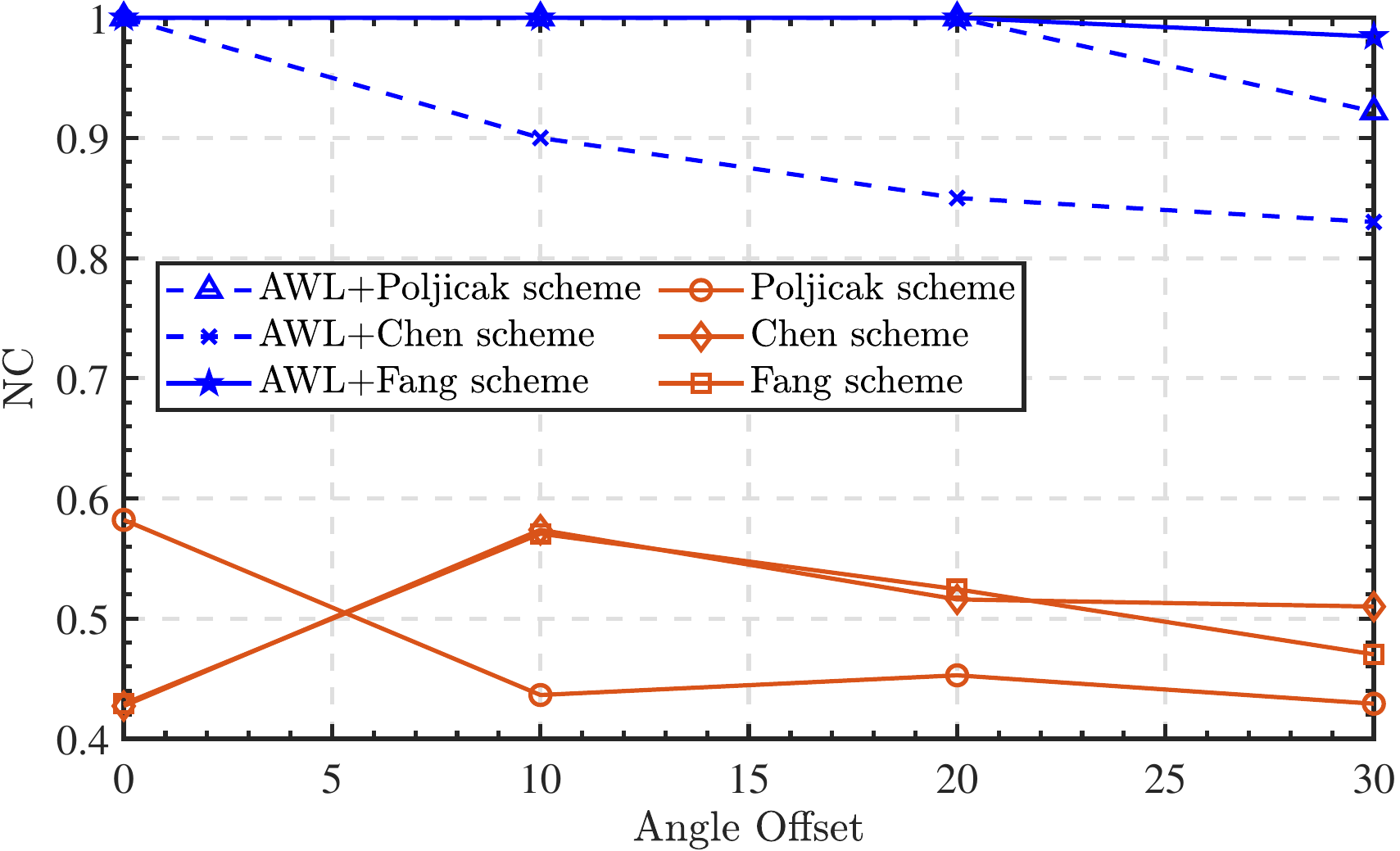}}
	\caption{The experimental results of robustness for AWD.}
	\label{fig:extexp}
\end{figure*}

The experimental results of robustness for AWD are shown in Fig.~\ref{fig:extexp}. After AWL and perspective correction, AWD without human assistance. We refer to Poljicak scheme \cite{poljicak2011discrete}, Chen scheme \cite{kang2010efficient} and Fang scheme \cite{fang2018screen} to embedding watermark, and the NC value of extracted message is as follow.
 \begin{equation}
NC=1-\frac{1}{P}\sum_{j=1}^P{|w^*\left( j \right) -w\left( j \right) |}
 \label{eq:Nc}
 \end{equation}
 where $w$,$w^*$, and $P$ are, respectively, the original watermark, the extracted watermark, and the size of the watermark.

Combined with proposing AWL algorithm, we can make these schemes detect watermarks automatically.

\subsection{Limitations}
The crux of detecting accuracy is the embedding intensity, and a higher embedding intensity guarantees better accuracy, but it inevitably decreases image transparency. Some marked images have conspicuous watermark regions, which can be easily detected by human eyes. However, during screen-shooting, the watermark regions may not be visible to camera devices, indicating that the embedding data is weaker than the noise introduced by the screen-shooting procedure. Thus, it might be a good idea to avoid excessive embedding strength by using a deep-learning-based approach to detect watermarking features in spatial domain.
\section{Conclusion}
In this paper, we present an intricately designed watermarking 
system to combat camera shooting. We propose an automatic watermark localization framework to accurately locate the region of interest (RoI), which contains watermark information,
in a screen-shooting image. In addition, we propose a blind detection algorithm to detect the RoI without any prior knowledge. To enhance the detection accuracy, we propose
a two-stage non-maximum suppression to obtain a more precise target region. Moreover, this framework is applicable to robust image watermarking that resists camera shooting. The experimental results show excellent detection accuracy in various complex scenarios, and we can automatically detect the watermark without human intervention by combining existing
methods.
\bibliographystyle{IEEEtran}
\bibliography{manuscripts}

\end{document}